\documentclass[11pt]{article}

\usepackage[final]{acl}

\usepackage{times}
\usepackage{latexsym}
\usepackage{xcolor}
\usepackage{colortbl}
\usepackage[T1]{fontenc}

\usepackage[utf8]{inputenc}

\usepackage{microtype}
\usepackage{threeparttable}

\usepackage{inconsolata}

\usepackage{graphicx}
\usepackage{multirow}
\usepackage{booktabs}

\usepackage[dvipsnames]{xcolor}
\newcommand{\best}[1]{\colorbox{purple!30}{#1}}    
\newcommand{\secondbest}[1]{\colorbox{orange!25}{#1}} 
\usepackage{tcolorbox}
\tcbuselibrary{breakable}
\usepackage{listings}
\newcommand{\prompttable}[4]{%
	\begin{table*}[p]
		\centering
		\begin{tcolorbox}[
			width=\textwidth,
			colback=gray!10,
			colframe=gray!75!black,
			title={#1}
			]
			\lstinputlisting[
			basicstyle=\rmfamily\small,
			breaklines=true,
			breakatwhitespace=false,
			breakautoindent=false,
			breakindent=0pt,
			columns=fullflexible,
			keepspaces=false,
			escapeinside={(*@}{@*)},
			showstringspaces=false
			]{#2}
		\end{tcolorbox}
		\caption{#3}
		\label{#4}
	\end{table*}
}

\definecolor{softpurple}{HTML}{E6CEE3}
\definecolor{softblue}{HTML}{E1E1F9}
\definecolor{softorange}{HTML}{FCD5B5}

\title{LifeMem: Enabling Lifelong Experience Reuse for LLM Agents}

\author{
	\textbf{Yuli Qiu\textsuperscript{1}},
	\textbf{Yutong Li\textsuperscript{1}},
	\textbf{Wei Su\textsuperscript{1}},
	\textbf{Zeming Liu\textsuperscript{2}},
	\textbf{Wanxiang Che\textsuperscript{3}},\\
	\textbf{Heyan Huang\textsuperscript{1}},
	\textbf{Haifeng Wang\textsuperscript{4}},
	\textbf{Yuhang Guo\textsuperscript{1, \textdagger}}
	\\
	\textsuperscript{1}School of Computer Science and Technology, Beijing Institute of Technology
	\\
	\textsuperscript{2}School of Computer Science and Engineering, Beihang University
	\\
	\textsuperscript{3}Research Center for Social Computing and Interactive Robotics, Harbin Institute of Technology
	\\
	\textsuperscript{4}Baidu Inc.
	\\
	\textsuperscript{\textdagger}Corresponding author
	\quad
	Email: 
	\href{mailto:yuliqiu@bit.edu.cn}{\textcolor{black}{yuliqiu}}@bit.edu.cn
	\href{mailto:guoyuhang@bit.edu.cn}{\textcolor{black}{guoyuhang}}@bit.edu.cn
}

\begin{document}
	\maketitle
	\begin{abstract}
		Large language model agents are expected to continuously adapt to new tasks and environments over their lifetime by reusing past experience. However, existing memory-based agents struggle to transfer reusable experience across environments and suffer from catastrophic forgetting as experience accumulated. To address these challenges, we propose LifeMem, a lifelong learning framework that enables agents to transfer knowledge across multiple environments. During learning, LifeMem clusters accumulated interaction trajectories based on underlying workflows to extract reusable skills. When solving a new task at inference time, the agent recalls relevant skills and trajectories to guide actions. To validate our method, we conduct experiments across 10 environments and over 13k tasks with 2k newly annotated interaction trajectories. Results show that LifeMem enables effective experience reuse in lifelong learning, achieving both reduced forgetting on learned tasks and superior cross-task transfer. Further analysis reveals that task streaming impacts learning, while consolidating structurally similar trajectories within memory boosts performance. Our dataset and code are available at \url{https://github.com/BITHLP/LifeMem}.
	\end{abstract}
	
	\section{Introduction}
	The pursuit of artificial general intelligence (AGI) requires autonomous agents that can continuously adapt to diverse and evolving environments over their lifetime \cite{xi2025rise, jiang2025adaptation}. Accordingly, large language model (LLM)-based agents are expected to generalize across multiple domains, such as embodied control, web navigation, and tool use, without relying on environment-specific redesigns \cite{zheng2025lifelong, wang2024agent}. To achieve this, agents have to learn reusable experience from new interaction trajectories without forgetting previously acquired experience \cite{cheng-etal-2026-mem2evolve, shan2026learning}, a capability known as \textbf{lifelong learning} \cite{zheng2025lifelongagentbench, zheng2025lifelong}. In particular, under the \textbf{inter-environment setting}, interaction trajectories are collected from heterogeneous environments with distinct action and observation spaces, which further complicates effective experience reuse \cite{zheng2025lifelong}.
	
	\begin{figure}[t]
		\includegraphics[width=\columnwidth]{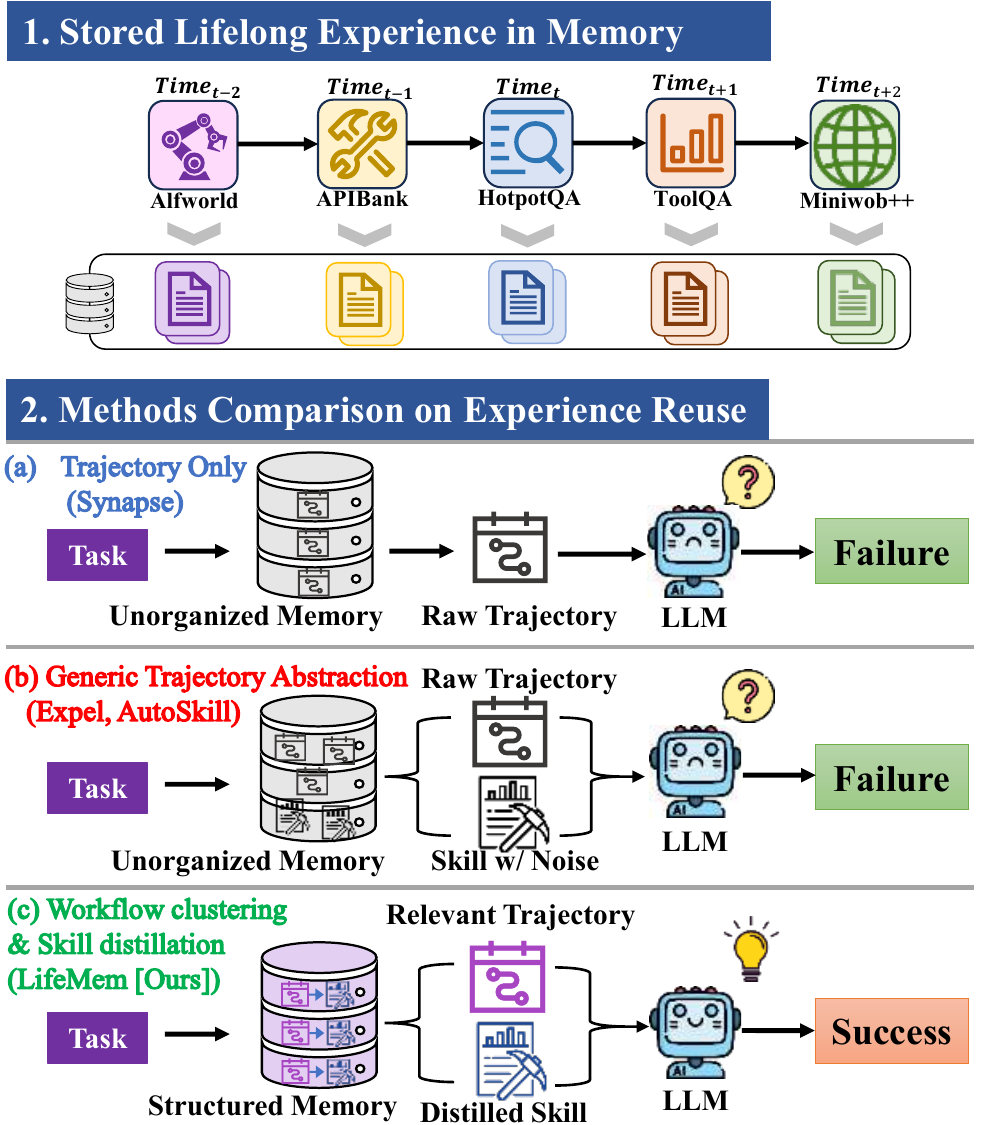}
		\caption{In lifelong learning, agent stores experience in memory. LifeMem enables effective experience reuse compared to previous methods.}
		\label{fig1}
	\end{figure}
	
	However, existing LLM-based agents still struggle to effectively reuse past experience to adapt to new environments. \textit{Internal-memory-based methods} \cite{yaoretroformer,qiao2024learn} fine-tune the model on trajectories from predefined environments, which restricts generalization and incurs high training costs \cite{tomilin2025meal, zhao2024expel}. \textit{External-memory-based methods} maintain an experience database and retrieve relevant trajectories to guide in-context learning \cite{xiang2024retrospex}, yet they encounter challenges as experience accumulates across heterogeneous environments \cite{zhao-etal-2024-sapt}: \textbf{(1) trajectory-only retrieval} \cite{zhengsynapse},  relying solely on raw trajectories, fails to bridge mismatched action spaces, causing weak generalization. Furthermore, as the database grows, it impairs accurate retrieval, leading to catastrophic forgetting of earlier environments. \textbf{(2) generic trajectory abstraction} \cite{zhao2024expel, Yang2026AutoSkillEL}, distilling high-level skills directly over an unorganized memory pool, allows irrelevant trajectories to introduce severe cross-environment noise. These noisy and unrepresentative skills are too coarse to guide novel tasks, leading to performance degradation (Figure~\ref{fig1}).
	
	To address these challenges, we propose \textbf{LifeMem}, a lifelong learning framework that enables agents to accumulate and reuse experience across multiple environments. LifeMem comprises two key components: \textbf{(1) workflow clustering,} which partitions trajectories based on underlying action workflows rather than surface semantic similarity, avoiding cross-environment retrieval interference. \textbf{(2) skill distillation,} instead of global abstraction, it distills skills locally from each workflow cluster, yielding highly representative skills untainted by noise. During \textbf{inference}, the agent leverages both environment-aligned trajectories and associated skills to support effective experience reuse without action-space interference.
	
	To validate the effectiveness of our method, we conduct experiments on 5 widely used agent scenarios \cite{chang2024agentboard, liuagentbench}: embodied action, tool utilization, web browsing, web search, and data analysis, covering 10 environments and more than 13k tasks with interaction trajectories \cite{yao2022react}. Notably, for 4 environments that lack interaction trajectories in existing benchmarks, we additionally annotate over 2k trajectories as training data. Experiment results show that LifeMem demonstrates superior lifelong learning performance over existing methods, characterized by two strengths: reduced forgetting on previously learned tasks, and more effective cross-task transfer to unseen domains. Further analysis indicates that the task streaming affects performance, where consolidating structurally similar tasks facilitates agent efficacy.
	
	Our contributions can be summarized as follows:
	\begin{itemize}
		\item We propose \textbf{LifeMem}, a novel lifelong learning framework for experience reuse across heterogeneous environments. By clustering trajectories into structured skills, LifeMem mitigates catastrophic forgetting while enabling cross-task transfer to unseen domains.
		
		\item To address the lack of agent-environment interaction trajectories in existing 4 environments, we contribute over 2k newly annotated trajectories as training data to enable agent learning.
		
		\item Through experimental analysis, we find that task streaming order impacts learning outcomes, and consolidating structurally similar trajectories within memory benefits agent performance.
	\end{itemize}
	
	\section{Related Work}
	\subsection{Lifelong Learning in LLMs}
	Lifelong learning enables LLMs to learn continuously and adaptively over their operational lifetime by integrating new knowledge while retaining previously acquired information, thereby mitigating catastrophic forgetting \cite{zheng2025towards}. (1) work on LLM lifelong learning primarily focuses on continual pretraining \cite{cossu2024continual, gupta2023continual}, fine-tuning \cite{sunlamol, huang2021continual}, and alignment  \cite{lin2024mitigating}, which are conducted on a sequence of predefined static datasets. (2) work on LLM agents lifelong learning mainly investigate experience accumulation within a single environment \cite{wang2024agent, zhao2024expel}, in which agents improve performance through repeated interactions under fixed state and action spaces.
	
	Different from prior work, we focus on lifelong experience accumulation for agents across multiple environments \cite{zheng2025lifelong}. In this setting, agents continuously integrate interaction trajectories originating from heterogeneous environments and learn to transfer experience across domains while preserving previously encountered ones.
	
	\subsection{Memory Mechanism of LLM Agents}
	
	LLM agents rely on memory mechanisms to manage and reuse past experience because of the limited context window \cite{yu2025memagent, chhikara2025mem0}. To this end, researchers adopt external vector database that store interaction trajectories and retrieve relevant ones based on semantic similarity to support in-context learning \cite{packer2023memgpt, liu2023think, wangvoyager, zhengsynapse}. However, they fail to transfer experience to tasks with different environments or other out-of-distribution settings \cite{zhang2025survey}.
	
	More recent works explore abstract memory representations to guide agent planning. However, frameworks utilizing dialogue summaries or knowledge graphs \cite{chhikara2025mem0, rasmussen2025zep} primarily target conversation compression rather than workflow reuse. Though advanced methods like AutoSkill \cite{Yang2026AutoSkillEL} and Hermes\footnote{https://github.com/NousResearch/hermes-agent} attempt to distill high-level skills directly from past trajectories, their global extraction over an unorganized experience pool allows irrelevant trajectories to act as cross-environment noise, severely degrading the representativeness of each skill. In contrast, our approach clusters trajectories into structural workflows before distillation, yielding highly representative skills.

	\section{Preliminaries}
	We consider a LLM agent interacting with an observable environment $\mathcal{E} = (S, A, \Omega, T)$, where $S$ denotes the environment states, $A$ the action space, $\Omega$ the observation space, and $T$ the transition dynamics. At each time step $t$, the agent receives an observation $o_t \in \Omega$ and selects an action
	\begin{equation}
		a_t = P_\theta(g, M, o_t)
	\end{equation}
	where $P_\theta$ is the language model policy, $g$ is a natural language goal, and $M$ is memory. The interaction continues until task termination.
	
	A task is defined as $\mathcal{T}^{i} = \langle \mathcal{E}^{i}, o^{i}_0, g^{i} \rangle$, 
	where $\mathcal{E}^{i}$ denotes the environment of the $i$-th task, $o^{i}_0$ is the initial observation, and $g^{i}$ is the task goal. Solving a task produces a trajectory $\xi^{i} = \{(o_0, a_0), (o_1, a_1), \ldots, (o_T, a_T)\}$.
	
	In lifelong learning, the agent encounters a sequence of tasks $\{\mathcal{T}^{1}, \mathcal{T}^{2}, \ldots, \mathcal{T}^{N}\}$. In the inter-environment setting, tasks arise from distinct environments $\mathcal{E}$. Trajectories collected from past tasks are incorporated into memory, as 
	\begin{equation}
		M^{i}=M^{i-1} \cup \{\xi^{i}\}
	\end{equation}
	to guide future task solving. The objective is to leverage accumulated experience to improve performance on new tasks while mitigating forgetting of previously learned behaviors.
	
	\section{Method}
	\begin{figure*}[t]
		\includegraphics[width=\textwidth]{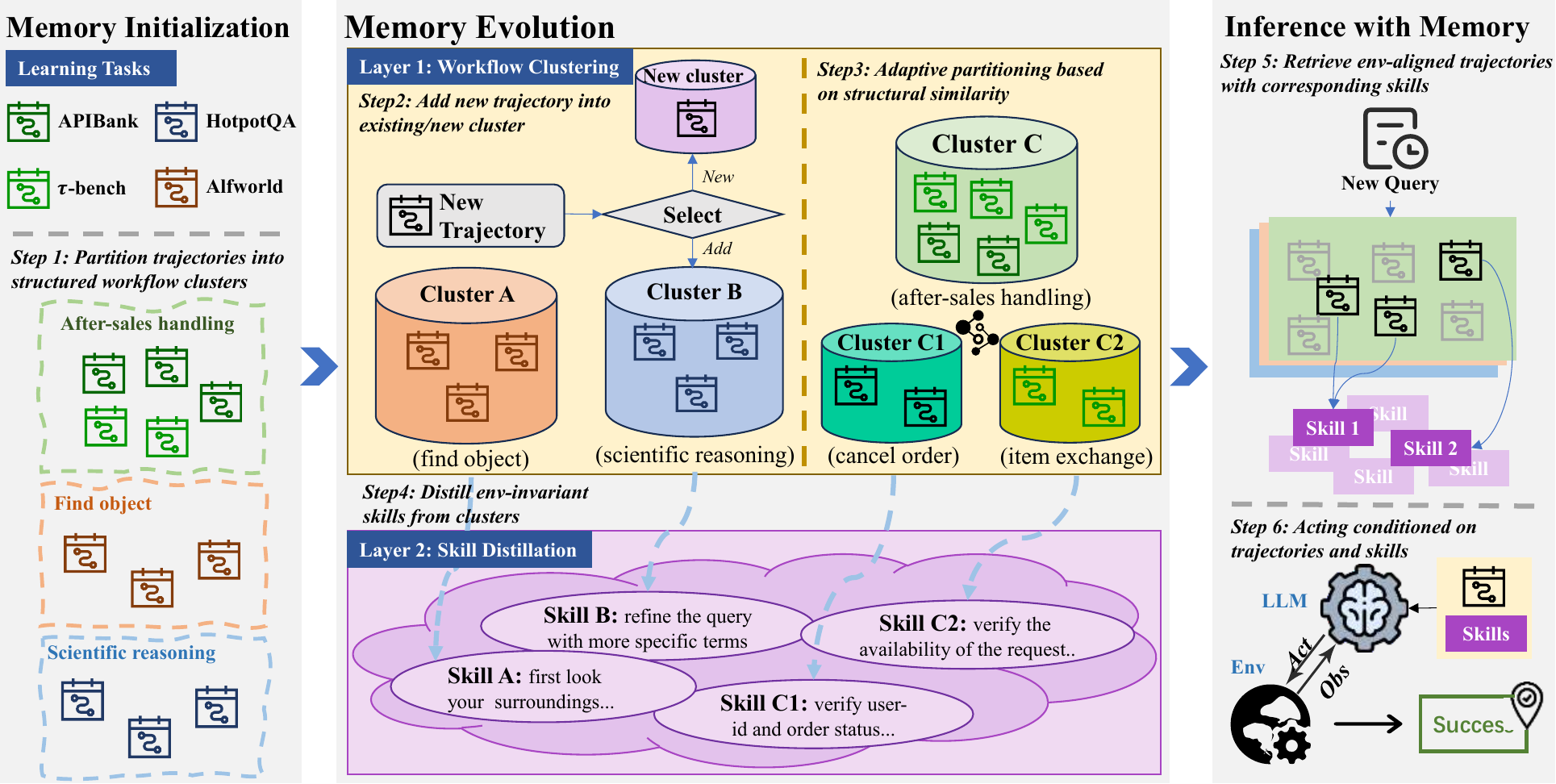}
		\caption{Overview of the LifeMem framework. In \textit{memory initialization}, LifeMem groups interaction trajectories from heterogeneous environments into structured workflow clusters. In \textit{memory evolution}, incoming trajectories are aligned with existing workflows or initialize a new one; as experience accumulates, workflow clusters are adaptively partitioned into finer-grained ones, from which transferable skills are distilled. In \textit{inference}, LifeMem activates environment-aligned trajectories along with their skills to guide the agent’s actions.}
		\label{fig2}
	\end{figure*}
	
	To enable experience accumulation and reuse in inter-environment lifelong learning, we model memory in an LLM agent as a dynamic component that is initialized once, incrementally updated with new experience, and queried during inference.
	
	\subsection{Memory Initialization}
	We initialize memory as a set of workflow-level clusters, which constitute the basic units for experience storage. Each cluster corresponds to a distinct behavioral workflow abstracted from trajectories.
	
	Given an initial set of trajectories $\{\xi^{0}, \ldots, \xi^{N}\}$, LLM partitions them into a set of clusters $C = \{c_1, \ldots, c_M\}$ by identifying common workflow patterns. For each cluster $c_i \in C$, a concise description $d_i$ is generated from its constituent trajectories, forming the initial memory.
	
	\subsection{Memory Evolution}
	\textbf{Workflow Clustering:} Since trajectories from different environments $\mathcal{E}$ may differ in goal semantics and action space while sharing underlying workflows, organizing them solely by semantic similarity is insufficient. We therefore update memory by assigning new trajectories to memory based on functional workflow alignment.
	
	When a new trajectory $\xi$ arrives, we retrieve its top-$k$ related trajectories from memory and collect their associated clusters $C_{\xi}$. Conditioned on $\xi$ and $C_{\xi}$, language model determines one of two actions: (1) if $\xi$ aligns with an existing cluster $c \in C_{\xi}$, it's added to $c$ and the corresponding description $d_c$ is updated. (2) if $\xi$ exhibits misalignment with $C_{\xi}$, a new cluster is created to accommodate it.
	
	As experience accumulates, a workflow cluster may begin to encompass heterogeneous behaviors, thereby reducing specificity. To mitigate this issue, we introduce a geometric diversity constraint based on the Mean Resultant Length (MRL) from directional statistics \cite{mardia2000directional}. Given a cluster $c$ containing $n$ trajectories, we first normalize each trajectory's embedding representation to a unit vector  $\mathbf{x}_i$ where $\|\mathbf{x}_i\| = 1$. The directional concentration $R$ is defined as:
	\begin{equation}
		R = \left| \frac{1}{n} \sum_{i=1}^{n} \mathbf{x}_i \right|
	\end{equation}
	The value $R \in [0, 1]$ directly quantifies structural homogeneity; $R \to 1$ implies high redundancy (e.g., repeating identical templates), while a lower $R$ indicates diverse operational behaviors. Accordingly, $c$ is partitioned into finer-grained sub-clusters if and only if both the volume threshold and diversity constraint are satisfied: $n > L$ and $R < \tau_{mrl}$. This enables LifeMem to adaptively adjust workflow granularity based on trajectory similarity, ensuring partitioning occurs only under genuine behavior divergence.
	
	\textbf{Skill Distillation:} Storing experience solely at the trajectory level limits continuous transfer, as trajectories collected in one environment seldom generalize to others due to mismatched observations or action spaces. To enable robust workflow reuse across heterogeneous tasks, we introduce a high-level skill abstraction layer, which distills each structured workflow cluster into concise, environment-invariant textual skills to guide subsequent agent execution.
	
	Formally, for each workflow cluster $c_i \in C$, we map it to a corresponding high-level skill:
	\begin{equation}
		s_i = f(c_i),
	\end{equation}
	where $f(\cdot)$ denotes the language model prompting process used to abstract trajectories contained in $c_i$. Each distilled skill $s_i$ captures the shared constraints, logical dependencies, and generalized execution principles decoupled from low-level environment details. The resulting hierarchical memory is formally denoted as:
	\begin{equation}
		\mathcal{H} = \{(c_i, s_i) \mid c_i \in C\}.
	\end{equation}
	
	\subsection{Inference with Memory}
	During inference, the agent must leverage past experience to solve the current task while avoiding interference from environment-specific behaviors. To this end, a dual-tier mechanism activates both environment-specific trajectories and environment-invariant skills to guide the agent.
	
	Given a target task, we first retrieve the top-$k$ most similar historical trajectories $\{\xi^{0}, \xi^{1}, \ldots, \xi^{k-1}\}$ from the memory. These trajectories are conditioned on the same environment $\mathcal{E}$ to provide explicit, environment-aligned operational exemplars for task execution.
	
	In parallel, we identify the workflow clusters $C_\xi$ associated with these retrieved trajectories and activate their corresponding skills:
	\begin{equation}
		S_{\xi} = \{s_i \mid c_i \in C_\xi\}
	\end{equation}
	Since trajectories from diverse environments sharing similar execution logic are grouped into the same clusters, this skill layer facilitates robust cross-environment transfer without introducing low-level action interference.
	
	Ultimately, the agent generates actions conditioned on both the concrete exemplars and the high-level skills:
	\begin{equation}
		a = P_\theta(g, M_{act}, o)
	\end{equation}
	where $M_{act} = ( \{\xi^{j}\}_{j=0}^{k-1}, S_{\xi} )$ denotes the activated memory contexts. This design allows LifeMem to benefit simultaneously from precise behavioral grounding and transferable skill abstractions.

	\section{Experiment}
	\subsection{Dataset Construction}
	
	\begin{table}[]
		\begin{threeparttable}
			{\setlength{\tabcolsep}{4pt}
				\fontsize{10}{12}\selectfont
				\begin{tabular}{llcc}
					\hline
					\textbf{Category}                                                                    & \textbf{Datasets}                                                      & \textbf{Trainset} & \textbf{Testset} \\ \hline
					\multirow{2}{*}{\begin{tabular}[c]{@{}l@{}}Embodied\\ Action\end{tabular}}  & \begin{tabular}[c]{@{}l@{}}AlfWorld\\ \cite{shridharalfworld}\end{tabular}          & 2204     & 134     \\ 
					& \begin{tabular}[c]{@{}l@{}}ScienceWorld \tnote{*}\\ \cite{wang2022scienceworld}\end{tabular}      & 2043     & 139     \\ \hline
					\multirow{2}{*}{\begin{tabular}[c]{@{}l@{}}Tool\\ Utilization\end{tabular}} & \begin{tabular}[c]{@{}l@{}}APIBank\\ \cite{li2023api}\end{tabular}           & 3131     & 262     \\
					& \begin{tabular}[c]{@{}l@{}}$\tau$-bench-airline\\ \cite{yao2024tau}\end{tabular}         & 530      & 115     \\ \hline
					\multirow{2}{*}{\begin{tabular}[c]{@{}l@{}}Web\\ Search\end{tabular}}       & \begin{tabular}[c]{@{}l@{}}HotpotQA\\ \cite{yang2018hotpotqa}\end{tabular}          & 800      & 100     \\
					& \begin{tabular}[c]{@{}l@{}}Webshop\\ \cite{yao2022webshop}\end{tabular}           & 1571     & 100     \\ \hline
					\multirow{2}{*}{\begin{tabular}[c]{@{}l@{}}Data \\ Analysis\end{tabular}}   & \begin{tabular}[c]{@{}l@{}}LifelongAgentBench\\ \cite{zheng2025lifelongagentbench}\end{tabular} & 280      & -      \\
					& \begin{tabular}[c]{@{}l@{}}ToolQA-coffee\\ \cite{zhuang2023toolqa}\end{tabular}            & 700      & 80      \\ \hline
					\multirow{2}{*}{\begin{tabular}[c]{@{}l@{}}Web\\ Browser\end{tabular}}      & \begin{tabular}[c]{@{}l@{}}Miniwob++\\ \cite{liu2018reinforcement}\end{tabular}         & 221      & 48      \\
					& \begin{tabular}[c]{@{}l@{}}Mind2web\\ \cite{deng2023mind2web}\end{tabular}          & 1009     & 252     \\ \hline
					\multicolumn{2}{l}{\textbf{Sum.}}                                                                                                                    &  12489        &   1230   \\ \hline
			\end{tabular}}
			\begin{tablenotes}
				\footnotesize
				\item[*] We perform stratified sampling from ScienceWorld to ensure the proportions of different task topics.
			\end{tablenotes}
		\end{threeparttable}
		\caption{Overview of tasks and datasets used in our lifelong learning setup.}
		\label{dataset}
	\end{table}
	
	\textbf{Overview:} To evaluate the ability of agents to reuse experience across heterogeneous environments, we conduct experiments on 5 widely used agent scenarios following prior work \cite{chang2024agentboard, liuagentbench}: embodied action \cite{shridharalfworld}, tool utilization \cite{li2023api}, web search \cite{yang2018hotpotqa}, data analysis \cite{zheng2025lifelongagentbench}, and web browsing \cite{liu2018reinforcement}. Specifically, we select ten tasks from diverse environments, comprising 12,489 training instances and 1,230 test instances (Table~\ref{dataset}). Details of datasets and environments are in Appendix~\ref{appendix:a}.
	
	\textbf{Data generation:} Since $\tau$-bench \cite{yao2024tau}, HotpotQA \cite{yang2018hotpotqa}, LifelongAgentBench \cite{zheng2025lifelongagentbench}, and ToolQA \cite{zhuang2023toolqa} don't provide training splits, we construct training data by collecting agent–environment interaction trajectories on them. Specifically, we prompt \textit{GPT-4.1}\footnote{https://developers.openai.com/api/docs/models/gpt-4.1} to solve these tasks under ReAct framework \cite{yao2022react}, yielding 2,310 trajectories, among which 1,541 are successful according to environment feedback. 
	
	\textbf{Quality control:} We conduct human evaluation to assess the quality of generated trajectories (Appendix~\ref{appendix:b}). The results show that over 90\% of the model's actions are judged to be reasonable, indicating that the generated trajectories are of high quality and suitable for guiding model learning.
	
	\begin{table*}[]
		\centering
		\setlength{\tabcolsep}{2.8pt}
		\renewcommand{\arraystretch}{1.08}
		\resizebox{\textwidth}{!}{%
			\begin{tabular}{ll*{9}{c}}
				\hline
				\multirow{2}{*}{\textbf{Model}} & \multirow{2}{*}{\textbf{Task}} & \multicolumn{5}{c}{\textbf{Overall Performance $\uparrow$}} & \multicolumn{4}{c}{\textbf{Backward Transfer $\uparrow$}} \\ \cmidrule(lr){3-7} \cmidrule(lr){8-11}
				& & ReAct & Synapse & ExpeL & AutoSkill & LifeMem & Synapse & ExpeL & AutoSkill & LifeMem \\ \hline
				\multirow{6}{*}{\textbf{\textit{GPT-4o-mini}}} & Embodied Action   & 10.62 & 40.02 & 44.09 & 12.36 & 45.65 & -3.85 & +8.81 & -28.97 & +14.84 \\
				& Tool Utilization  & 35.18 & 30.93 & 36.87 & 34.40 & 37.93 & -11.33 & -1.63 & +2.84 & +0.42 \\
				& Web Search        & 32.50 & 31.50 & 33.50 & 29.50 & 39.50 & -9.89 & +4.80 & -6.65 & +6.91 \\
				& Data Analysis     & 73.53 & 68.75 & 65.00 & 45.00 & 68.75 & -3.51 & +0.53 & -2.70 & -1.79 \\
				& Web Browsing      & 14.73 & 39.49 & 41.37 & 50.72 & 41.15 & -2.70 & +1.36 & 0.00 & +2.77 \\ \cline{2-11}
				& \textbf{Avg.} & 28.84 & 39.18 & \secondbest{41.85} & 33.22 & \best{$\textbf{44.13}_{\uparrow 5.45\%}$} & -7.04 & \secondbest{+3.10} & -8.54 & \best{\textbf{+5.66}} \\ \hline
				\multirow{6}{*}{\textbf{\textit{Deepseek-v3.2-exp}}} & Embodied Action   & 38.98 & 52.89 & 35.28 & 39.27 & 54.95 & -3.17 & +3.29 & -17.03 & -1.82 \\
				& Tool Utilization  & 49.34 & 53.69 & 54.07 & 52.87 & 52.70 & -0.64 & -1.10 & +5.10 & -1.56 \\
				& Web Search        & 30.00 & 36.00 & 39.50 & 27.00 & 40.00 & -1.25 & 0.00 & -26.86 & +12.66 \\
				& Data Analysis     & 37.50 & 63.75 & 71.25 & 61.25 & 68.75 & +2.00 & -3.39 & 0.00 & +1.85 \\
				& Web Browsing      & 48.07 & 34.28 & 32.84 & 33.04 & 34.08 & +3.33 & -1.61 & +2.43 & +3.33 \\ \cline{2-11}
				& \textbf{Avg.} & 41.14 & \secondbest{46.38} & 43.85 & 40.62 & \best{$\textbf{48.02}_{\uparrow 3.54\%}$} & -0.60 & \secondbest{-0.28} & -9.09 & \best{\textbf{+3.11}} \\ \hline
				\multirow{6}{*}{\textbf{\textit{Qwen3-32b}}} & Embodied Action   & 28.09 & 52.72 & 55.62 & 35.41 & 55.61 & +1.65 & +3.52 & +7.11 & +9.27 \\
				& Tool Utilization  & 39.04 & 41.12 & 39.90 & 28.32 & 44.09 & -3.29 & -9.96 & -8.81 & +4.54 \\
				& Web Search        & 35.00 & 33.50 & 35.00 & 33.50 & 42.00 & -2.57 & +5.60 & -8.07 & 0.00 \\
				& Data Analysis     & 37.50 & 67.50 & 75.00 & 47.50 & 70.00 & -1.82 & 0.00 & -28.36 & +9.80 \\
				& Web Browsing      & 32.04 & 27.72 & 26.79 & 40.09 & 32.64 & -7.70 & -7.70 & -2.69 & +16.00 \\ \cline{2-11}
				& \textbf{Avg.} & 33.98 & 41.96 & \secondbest{43.29} & 35.79 & \best{$\textbf{46.52}_{\uparrow 7.46\%}$} & -2.24 & \secondbest{-1.17} & -6.17 & \best{\textbf{+6.68}} \\ \hline
			\end{tabular}%
		}
		\caption{Performance of various agent memory methods under inter-environment lifelong learning. For each category, the reported values are the averages over the datasets belonging to that category. \textbf{Avg.} denotes the average over all test datasets. The best and second-best results are marked \colorbox{purple!30}{\textbf{purple}} and \colorbox{orange!25}{orange}, respectively.}
		\label{main-result}
	\end{table*}
	
	\subsection{Baselines}
	We consider three baselines that represent different experience reuse strategies. ReAct \cite{yao2022react} serves as a \textbf{memory-free} baseline, which relies solely on step-by-step reasoning and action without past experience. In contrast, we adopt three representative \textbf{external-memory-based methods}: Synapse \cite{zhengsynapse} abstracts the current state and retrieves relevant trajectories for in-context learning, whereas ExpeL \cite{zhao2024expel} and AutoSkill \cite{Yang2026AutoSkillEL} shift towards extracting high-level experience from raw trajectories. Specifically, the former distills insights from trial and error, whereas the latter uncovers operational skills from successful rollouts. We use GPT-4.1 to extract experience for all baselines.
	
	We use three widely adopted models for experiments: GPT-4o-mini \cite{achiam2023gpt}, Deepseek-v3.2-exp \cite{liu2025deepseek}, and Qwen3-32b \cite{yang2025qwen3}. They come from different families, span both proprietary and open-source settings, and vary in parameter scale, providing a representative evaluation across LLM backbones.
	
	\subsection{Evaluation Metrics}
	Following prior work \cite{srinivasan2022climb, tomilin2025meal}, the evaluation of lifelong learning can be viewed from two stages: the extent to which previously learned tasks transfer experience to the current task, and the impact of subsequently learned tasks to the current. Let $s_{1,\ldots, j}^i$ denote the performance of task $i$ after trained on tasks $1, \ldots, j$, and let the number of training tasks be $N$, we assess the lifelong learning capability from following  perspectives.
	
	\textbf{Overall Performance:} We compute the average performance of the model across all tasks after learning all tasks \cite{tang2024vilco}:
	\begin{equation}
		\label{eq:example}
		OP = \frac{1}{N}\sum_{i=1}^{N}s_{1, \ldots, N}^i
	\end{equation}
	which provides an overall assessment perspective.
	
	
	\textbf{Backward Transfer:} By comparing the model’s performance on task $i$ after learning all tasks with that after learning only tasks $1, \ldots, i$, we measure the impact of subsequent tasks on performance of task $i$ \cite{tang2024vilco, srinivasan2022climb}:
	\begin{equation}
		\label{eq:example}
		BwT = \frac{1}{N-1}\sum_{i=1}^{N-1}\frac{s_{1, \ldots, N}^i-s_{1, \ldots, i}^i}{s_{1, \ldots, i}^i}
	\end{equation}
	a positive value indicates that subsequent tasks may facilitate earlier tasks, whereas a negative value reflects forgetting caused by subsequent tasks.
	
	We present these metrics in a more intuitive manner in Appendix~\ref{appendix:c}.
	
	\subsection{Implementation Details}
	In our memory implementation, we use SQLite\footnote{https://sqlite.org/index.html} to store all clusters and FAISS \cite{douze2024faiss} to index and retrieve trajectories. Following prior work \cite{lv2026all}, we utilize \textit{all-MiniLM-L6-v2} \cite{wang2021minilmv2} as our embedding model. During the initialization stage, the number of initial trajectories is set to 5. In the memory evolution, the retriever returns the top-4 relevant trajectories, and the cluster splitting threshold is set to $L=50$ and $\tau_{mrl}=0.95$. We employ a lightweight language model \textit{GPT-5-mini} \footnote{https://developers.openai.com/api/docs/models/gpt-5-mini} to efficiently determine the cluster assignment of incoming trajectories. For cluster splitting and skill abstraction, we use a more powerful model \textit{GPT-4.1}. Detailed prompts and hyper-parameters are in Appendix~\ref{appendix:d}.
	
	Overall, we organize the lifelong task stream to reflect a common non-stationary deployment dynamic where temporally adjacent tasks often exhibit semantic and operational locality. Specifically, tasks from homogeneous environments are sequentially grouped to simulate distinct operational phases (AlfWorld → ScienceWorld → APIBank → $\tau$-bench → HotpotQA → Webshop → LifelongAgentBench → ToolQA → MiniwoB++ → Mind2web). Within each category, tasks follow a coarse easy-to-hard progression to evaluate structural adaptation under structured data streams, adhering to standard curriculum learning benchmarks \cite{Wang2021ASO}. To further explore the system's robustness, we analyze the impact of fully randomized and buffer-based streams in Section~\ref{rq3}.
	
	\subsection{Main Results}
	\begin{figure}[h]
		\includegraphics[width=\columnwidth]{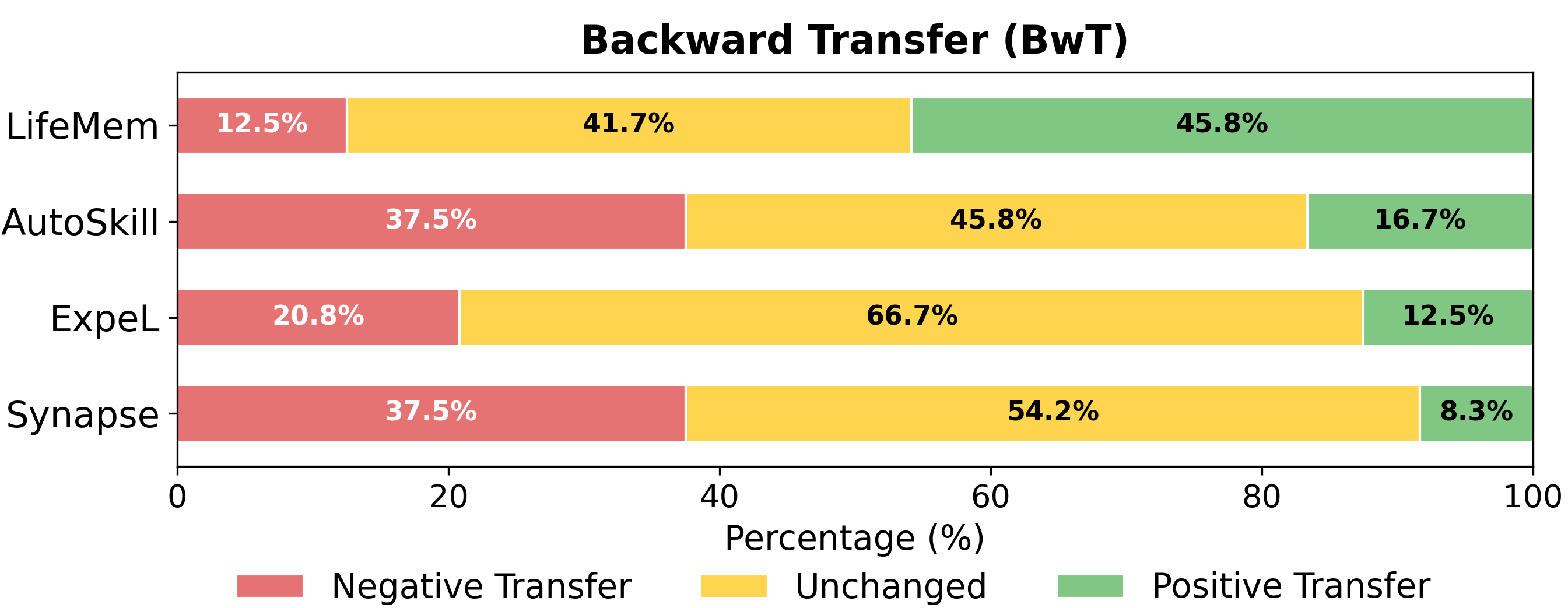}
		\caption{Distribution of transfer behaviors across all datasets during lifelong learning. Scores within the range of [-5, 5] are categorized as "Unchanged" representing cases with no obvious performance shift.}
		\label{fig3}
	\end{figure}
	
	\textbf{Overall Performance:} Table~\ref{main-result} reports the \textit{overall performance} and \textit{backward transfer} scores of different methods across five agent scenarios. LifeMem consistently outperforms all baselines across all metrics, demonstrating stronger overall lifelong learning performance. We further observe that experience-based learning yields larger performance gains for weaker backbone models. For example, GPT-4o-mini improves from 28.84 to 44.13, whereas the stronger model Deepseek-v3.2-exp shows a smaller increase from 41.14 to 48.02. This suggests that LifeMem benefits weaker models more substantially by distilling reusable skills from interaction trajectories to guide problem solving, thereby partially narrowing the performance gap across backbone models.
	
	\textbf{Transfer Behavior:} To evaluate how subsequent learning impacts past knowledge, we analyze both the average BwT values and the distribution of transfer events (Figure~\ref{fig3}). In baseline methods, as the number of learned tasks increases, the unstructured memory inevitably becomes cluttered. This leads to severe negative transfer due to inference interference: for the retrieval baseline Synapse, the cluttered memory causes the retrieval of irrelevant trajectories; the lack of experience isolation allows ongoing learning to overwrite and degrade the representativeness of early skills. 
	
	In contrast, LifeMem systematically structures the memory via explicit workflow clustering, effectively mitigating cross-task interference to achieve a superior, positive average BwT value in Table~\ref{main-result}. By organizing trajectories into distinct, structurally-aligned clusters, LifeMem significantly suppresses negative transfer events and maximizes positive transfer, allowing experience from subsequent tasks to consistently reinforce rather than interfere with earlier knowledge.

	\section{Analysis}
	In this section, we conduct a comprehensive analysis aimed at addressing the following research questions. \textbf{RQ1:} \textit{Which components of our method are critical to its performance in lifelong learning?} (\S \ref{rq1}) \textbf{RQ2:} \textit{How effective is LifeMem in facilitating zero-shot cross-task experience transfer?} (\S\ref{rq2}) \textbf{RQ3:} \textit{How do different task streaming strategies influence experience reuse performance?} (\S\ref{rq3}) \textbf{RQ4:} \textit{Can boosting retrieval components directly bridge the performance gap between baselines and LifeMem?} (\S\ref{retrieval_study}) \textbf{RQ5:} \textit{What is the impact of hyperparameters?} (\S\ref{hyper})
	
	\subsection{RQ1: Ablation Study}
	\label{rq1}
	\begin{table}[htbp]
		\centering
		\label{tab:ablation_new}
		\resizebox{\columnwidth}{!}{%
			\begin{tabular}{lc} \toprule
				\textbf{Method} & \textbf{Overall Performance} \\
				\hline \rowcolor{gray!20}
				ReAct (No memory) & \\ \hline
				\quad \textit{gpt-4o-mini} & 28.84 \\
				\quad \textit{gpt-4.1} & 40.42 \\
				\hline \rowcolor{gray!20}
				LifeMem & \\ \hline
				\quad \textit{+trajs} & 42.84 \\
				\quad \textit{+skills (by gpt-4o-mini)} & 36.16 \\
				\quad \textit{+skills (by gpt-4.1)} & 37.61 \\
				\quad \textit{+trajs+skills (by gpt-4o-mini)} & 43.18 \\
				\quad \textit{+trajs+skills (by gpt-4.1)} & \textbf{44.13} \\
				\bottomrule
			\end{tabular}%
		}
	\caption{Component-wise ablation study on LifeMem. We evaluate the incremental effects of our env-aligned trajectories (\textit{trajs}) and abstracted skills (\textit{skills}) across gpt-4o-mini as backbone. The model in parentheses indicates the generator used to extract the skills.}
	\label{ablation1}
	\end{table}
	
	To dissect the contribution of each individual component within LifeMem, we conduct a multi-dimensional ablation study. Specifically, we evaluate the system performance under three configuration granularities: providing historical trajectories alone, incorporating abstracted workflows, and utilizing both components simultaneously. 
	
	Our ablation results in Table~\ref{ablation1} reveal that both environment-aligned trajectories and abstracted skills consistently provide positive contributions to LifeMem. First, the inclusion of \textit{trajs} provides raw demonstrations for the agent to mimic successful task execution formats, which is foundational for interactive agent tasks. Second, the abstracted \textit{skills} further elevate the performance by offering macro-guidance, with the joint configuration yielding the optimal success rate. 
	
	Furthermore, we observe that the capacity of the skill generator plays a nuanced role. A more powerful gpt-4.1 exhibits a marginal advantage over gpt-4o-mini in extracting high-level skills from trajectories. Crucially, even when LifeMem relies entirely on gpt-4o-mini to extract skills, it still delivers a substantial gain over the baseline. Besides, LifeMem with a gpt-4o-mini backbone substantially outperforms the vanilla performance of gpt-4.1, verifying that the core efficacy of LifeMem stems from extracting and reusing skills from past trajectories, rather than distilling a superior LLM. We further present the ablation results of other models in Appendix~\ref{ablation}.
	
	\subsection{RQ2: Zero-Shot Cross-Task Transfer}
	\label{rq2}
	\begin{table}[htbp]
		\centering
		\resizebox{\columnwidth}{!}{%
			\begin{tabular}{lcc}
				\hline 
				\textbf{Method} & GPT-4o-mini & Deepseek-v3.2-exp \\ 
				\hline 
				\rowcolor{gray!20} \multicolumn{3}{c}{\textit{FEVER}} \\ 
				ReAct & 0.40 & 0.39 \\
				Synapse & 0.38 & 0.43 \\
				ExpeL & 0.41 & 0.32 \\ 
				AutoSkill & \textbf{0.49} & 0.36 \\ \hline
				LifeMem & \textbf{0.49}\rlap{$_{\uparrow 22.5\%}$} & \textbf{0.46}\rlap{$_{\uparrow 17.9\%}$} \\ 
				LifeMem \textit{(w/o trajs)} & 0.47\rlap{$_{\uparrow 17.5\%}$} & 0.43\rlap{$_{\uparrow 10.3\%}$} \\ 
				\hline 
				\rowcolor{gray!20} \multicolumn{3}{c}{\textit{$\tau$-bench (airline)}} \\ 
				ReAct & 0.32 & 0.38 \\
				Synapse & 0.26 & 0.40 \\
				ExpeL & 0.26 & 0.42 \\
				AutoSkill & 0.34 & 0.36 \\ \hline
				LifeMem & \textbf{0.38}\rlap{$_{\uparrow 18.8\%}$} & \textbf{0.44}\rlap{$_{\uparrow 15.8\%}$} \\ 
				LifeMem \textit{(w/o trajs)} & \textbf{0.38}\rlap{$_{\uparrow 18.8\%}$} & 0.42\rlap{$_{\uparrow 10.5\%}$} \\ 
				\hline
			\end{tabular}%
		}
		\caption{Performance of zero-shot experience transfer to unseen tasks. Percentages represent the cross-task transfer improvements compared to ReAct. \textit{w/o trajs} means setting that only abstracted skills are provided.}
		\label{ood1}
	\end{table}
	To evaluate the zero-shot cross-task transfer capability of LifeMem, we investigate the following scenarios: (1) \textit{transfer to unseen task formulations} within an encountered environment (evaluating on the fact verification task FEVER \cite{thorne2018fever} using the same Wikipedia environment from HotpotQA), and (2) \textit{transfer to unseen environments} with a similar task formulation (evaluating on the airline task from $\tau$-bench \cite{yao2024tau} against the retail task). These configurations rigorously test zero-shot transfer, forcing the agent to adapt accumulated experience to unfamiliar domains without prior task-specific training.
	
	As shown in Table~\ref{ood1}, baselines achieve limited gains or even suffer from performance degradation compared to ReAct. For the first setting (FEVER), the semantic shift in task formulation causes raw-retrieval methods (Synapse and ExpeL) to retrieve irrelevant trajectories that misguide actions, while AutoSkill fails to match any compliant skill cluster for the unseen formulation. For the second setting, relying solely on raw trajectories fails to bridge the domain gap, as environmental discrepancies severely hinder direct experience reuse (See Appendix~\ref{appendix:h} for case study).
	
	In contrast, LifeMem delivers strong zero-shot transfer by successfully leveraging environment-invariant skills that capture the domain-agnostic logic of task execution. Consequently, even when facing unseen task formulations or unfamiliar environments, LifeMem effectively transfers these high-level skills to guide agent actions, demonstrating superior flexibility in open-ended target domains.
	
	\subsection{RQ3: Impact of Task Stream Strategies}
	\label{rq3}
	\begin{table}[h]
		\centering
		\resizebox{\columnwidth}{!}{%
		\begin{tabular}{lcc}
			\hline
			\textbf{Task Stream} & GPT-4o-mini & Deepseek-v3.2-exp \\ \hline
			Similarity & 44.13 & 48.02 \\
			Random & 39.63 & 44.34 \\
			Buffer-based & 41.64 \textcolor{ForestGreen}{\textit{\rlap{$_{\uparrow 44.7\%}$}}} & 46.73 \textcolor{ForestGreen}{\rlap{$_{\uparrow 64.9\%}$}}\\ \hline
		\end{tabular}}
		\caption{Robustness analysis of LifeMem across different task streaming strategies. \textcolor{ForestGreen}{Percentages} denote the performance recovery rate relative to the degradation under the random stream.}
		\label{order}
	\end{table}
	
	Task stream sequence effects learning performance \cite{tang2024vilco}. We investigate three different streams: \textbf{similarity}, sequential learning of tasks from identical scenarios; \textbf{random}, which uses a completely shuffled instance sequence, and \textbf{buffer-based}, where incoming random trajectories are temporarily held in a minor task-specific cache (size=5) and committed as a consolidated mini-batch to stabilize the unstructured stream.
	
	As reported in Table~\ref{order}, grouping related tasks improves memory stability and reduces retrieval interference under streaming experience accumulation. Crucially, even when encountering uncurated real-world streams, LifeMem can be easily optimized via the buffer-based strategy, enabling seamless stabilization. More results in random settings are in Appendix~\ref{random-base}.
	
	\section{Conclusion}
	In this work, we propose LifeMem, a novel lifelong learning framework that structures interaction trajectories into abstracted skills to enable effective experience reuse across heterogeneous environments. Evaluated on 10 environments and over 13k tasks, LifeMem significantly improves lifelong performance, successfully mitigating forgetting on learned tasks while achieving robust zero-shot cross-task transfer. Our analysis highlights a clear synergy: environment-aligned trajectories effectively guide action formats, whereas abstracted skills generalize better to unfamiliar scenarios. Ultimately, these insights provide practical design guidelines for how to better manage accumulated memory and extract reusable experience, enabling continual evolution after agent deployment.

		\section*{Limitations}
	While our results establish the efficacy of LifeMem in integrating and reusing cross-environment experience during lifelong learning, fully unlocking its potential for large-scale industrial deployments introduces exciting opportunities for engineering-level co-optimization. For instance, transitioning this framework to real-world applications with millions of stored experiences could be further accelerated by deploying lightweight, on-device language models to ensure local efficiency, or incorporating approximate retrieval mechanisms alongside other system-level enhancements.  
        \section*{Ethical Consideration}
    All datasets used in this work are publicly available and released for research purposes. We strictly follow their original licenses and intended usage. For benchmarks without official training splits, we construct training data by collecting agent-environment interaction trajectories generated under controlled task-oriented environments. The collected data contains no Personally Identifiable Information (PII), sensitive content, or offensive material.
    
    We ensure that all employed human annotators underwent rigorous training specific to the task categories involved in the evaluated environments. All participants were fully informed of the intended data usage prior to their involvement and were compensated fairly. Human evaluation is used solely to assess the reasonableness of model-generated interaction steps, ensuring data quality while minimizing subjective annotation.
    
    \section*{Acknowledgments}
    
    We thank all the anonymous reviewers for their insightful and valuable comments. This work is supported by the National Natural Science Foundation
    of China (Grant No.U21B2009).		
		
		
		
		\bibliography{custom}

@inproceedings{zhao2024expel,
	title={Expel: Llm agents are experiential learners},
	author={Zhao, Andrew and Huang, Daniel and Xu, Quentin and Lin, Matthieu and Liu, Yong-Jin and Huang, Gao},
	booktitle={Proceedings of the AAAI Conference on Artificial Intelligence},
	volume={38},
	number={17},
	pages={19632--19642},
	year={2024}
}

@inproceedings{zhengsynapse,
	title={Synapse: Trajectory-as-Exemplar Prompting with Memory for Computer Control},
	author={Zheng, Longtao and Wang, Rundong and Wang, Xinrun and An, Bo},
	year={2024},
	booktitle={The Twelfth International Conference on Learning Representations}
}

@article{zheng2025towards,
	title={Towards lifelong learning of large language models: A survey},
	author={Zheng, Junhao and Qiu, Shengjie and Shi, Chengming and Ma, Qianli},
	journal={ACM Computing Surveys},
	volume={57},
	number={8},
	pages={1--35},
	year={2025},
	publisher={ACM New York, NY}
}

@article{zheng2025lifelong,
	title={Lifelong learning of large language model based agents: A roadmap},
	author={Zheng, Junhao and Shi, Chengming and Cai, Xidi and Li, Qiuke and Zhang, Duzhen and Li, Chenxing and Yu, Dong and Ma, Qianli},
	journal={arXiv preprint arXiv:2501.07278},
	year={2025}
}

@article{qiao2024learn,
	title={Learn more, but bother less: parameter efficient continual learning},
	author={Qiao, Fuli and Mahdavi, Mehrdad},
	journal={Advances in Neural Information Processing Systems},
	volume={37},
	pages={97476--97498},
	year={2024}
}

@article{wang2024agent,
	title={Agent workflow memory},
	author={Wang, Zora Zhiruo and Mao, Jiayuan and Fried, Daniel and Neubig, Graham},
	journal={arXiv preprint arXiv:2409.07429},
	year={2024}
}

@article{zheng2025lifelongagentbench,
	title={LifelongAgentBench: Evaluating LLM Agents as Lifelong Learners},
	author={Zheng, Junhao and Cai, Xidi and Li, Qiuke and Zhang, Duzhen and Li, ZhongZhi and Zhang, Yingying and Song, Le and Ma, Qianli},
	journal={arXiv preprint arXiv:2505.11942},
	year={2025}
}

@inproceedings{yaoretroformer,
	title={Retroformer: Retrospective Large Language Agents with Policy Gradient Optimization},
	author={Yao, Weiran and Heinecke, Shelby and Niebles, Juan Carlos and Liu, Zhiwei and Feng, Yihao and Xue, Le and RN, Rithesh and Chen, Zeyuan and Zhang, Jianguo and Arpit, Devansh and others},
	booktitle={The Twelfth International Conference on Learning Representations},
	year={2023}
}

@inproceedings{xiang2024retrospex,
	title={Retrospex: Language agent meets offline reinforcement learning critic},
	author={Xiang, Yufei and Shen, Yiqun and Zhang, Yeqin and Cam-Tu, Nguyen},
	booktitle={Proceedings of the 2024 Conference on Empirical Methods in Natural Language Processing},
	pages={4650--4666},
	year={2024}
}

@article{chang2024agentboard,
	title={Agentboard: An analytical evaluation board of multi-turn llm agents},
	author={Chang, Ma and Zhang, Junlei and Zhu, Zhihao and Yang, Cheng and Yang, Yujiu and Jin, Yaohui and Lan, Zhenzhong and Kong, Lingpeng and He, Junxian},
	journal={Advances in neural information processing systems},
	volume={37},
	pages={74325--74362},
	year={2024}
}

@inproceedings{liuagentbench,
	title={AgentBench: Evaluating LLMs as Agents},
	author={Liu, Xiao and Yu, Hao and Zhang, Hanchen and Xu, Yifan and Lei, Xuanyu and Lai, Hanyu and Gu, Yu and Ding, Hangliang and Men, Kaiwen and Yang, Kejuan and others},
	booktitle={The Twelfth International Conference on Learning Representations},
	year={2023}
}

@article{srinivasan2022climb,
	title={Climb: A continual learning benchmark for vision-and-language tasks},
	author={Srinivasan, Tejas and Chang, Ting-Yun and Pinto Alva, Leticia and Chochlakis, Georgios and Rostami, Mohammad and Thomason, Jesse},
	journal={Advances in Neural Information Processing Systems},
	volume={35},
	pages={29440--29453},
	year={2022}
}

@article{tang2024vilco,
	title={Vilco-bench: Video language continual learning benchmark},
	author={Tang, Tianqi and Deldari, Shohreh and Xue, Hao and De Melo, Celso and Salim, Flora},
	journal={Advances in Neural Information Processing Systems},
	volume={37},
	pages={70213--70229},
	year={2024}
}

@inproceedings{achiam2023gpt,
	title={GPT-4 Technical Report},
	author={OpenAI Josh Achiam and Steven Adler and Sandhini Agarwal and Lama Ahmad and Ilge Akkaya and Florencia Leoni Aleman and Diogo Moitinho de Almeida and Janko Altenschmidt and Sam Altman and Shyamal Anadkat and Red Avila and Igor Babuschkin and Suchir Balaji and Valerie Balcom and Paul Baltescu and Haim-ing Bao and Mo Bavarian and Jeff Belgum and Irwan Bello and Jake Berdine and Gabriel Bernadett-Shapiro and Christopher Berner and Lenny Bogdonoff and Oleg Boiko and Made-laine Boyd and Anna-Luisa Brakman and Greg Brockman and Tim Brooks and Miles Brundage and Kevin Button and Trevor Cai and Rosie Campbell and Andrew Cann and Brittany Carey and Chelsea Carlson and Rory Carmichael and Brooke Chan and Che Chang and Fotis Chantzis and Derek Chen and Sully Chen and Ruby Chen and Jason Chen and Mark Chen and Benjamin Chess and Chester Cho and Casey Chu and Hyung Won Chung and Dave Cummings and Jeremiah Currier and Yunxing Dai and Cory Decareaux and Thomas Degry and Noah Deutsch and Damien Deville and Arka Dhar and David Dohan and Steve Dowling and Sheila Dunning and Adrien Ecoffet and Atty Eleti and Tyna Eloundou and David Farhi and Liam Fedus and Niko Felix and Sim'on Posada Fishman and Juston Forte and Is-abella Fulford and Leo Gao and Elie Georges and Christian Gibson and Vik Goel and Tarun Gogineni and Gabriel Goh and Raphael Gontijo-Lopes and Jonathan Gordon and Morgan Grafstein and Scott Gray and Ryan Greene and Joshua Gross and Shixiang Shane Gu and Yufei Guo and Chris Hallacy and Jesse Han and Jeff Harris and Yuchen He and Mike Heaton and Johannes Heidecke and Chris Hesse and Alan Hickey and Wade Hickey and Peter Hoeschele and Brandon Houghton and Kenny Hsu and Shengli Hu and Xin Hu and Joost Huizinga and Shantanu Jain and Shawn Jain and Joanne Jang and Angela Jiang and Roger Jiang and Haozhun Jin and Denny Jin and Shino Jomoto and Billie Jonn and Heewoo Jun and Tomer Kaftan and Lukasz Kaiser and Ali Kamali and Ingmar Kanitscheider and Nitish Shirish Keskar and Tabarak Khan and Logan Kilpatrick and Jong Wook Kim and Christina Kim and Yongjik Kim and Hendrik Kirchner and Jamie Ryan Kiros and Matthew Knight and Daniel Kokotajlo and Lukasz Kondraciuk and Andrew Kondrich and Aris Konstantinidis and Kyle Kosic and Gretchen Krueger and Vishal Kuo and Michael Lampe and Ikai Lan and Teddy Lee and Jan Leike and Jade Leung and Daniel Levy and Chak Li and Rachel Lim and Molly Lin and Stephanie L. Lin and Ma-teusz Litwin and Theresa Lopez and Ryan Lowe and Patricia Lue and Anna Makanju and Kim Malfacini and Sam Manning and Todor Markov and Yaniv Markovski and Bianca Martin and Katie Mayer and Andrew Mayne and Bob McGrew and Scott Mayer McKinney and Christine McLeavey and Paul McMillan and Jake McNeil and David Medina and Aalok Mehta and Jacob Menick and Luke Metz and Andrey Mishchenko and Pamela Mishkin and Vinnie Monaco and Evan Morikawa and Daniel P. Mossing and Tong Mu and Mira Murati and Oleg Murk and David M'ely and Ashvin Nair and Reiichiro Nakano and Rajeev Nayak and Arvind Neelakantan and Richard Ngo and Hyeonwoo Noh and Ouyang Long and Cullen O'Keefe and Jakub W. Pachocki and Alex Paino and Joe Palermo and Ashley Pantuliano and Giambattista Parascandolo and Joel Parish and Emy Parparita and Alexandre Passos and Mikhail Pavlov and Andrew Peng and Adam Perelman and Filipe de Avila Belbute Peres and Michael Petrov and Henrique Pond{\'e} de Oliveira Pinto and Michael Pokorny and Michelle Pokrass and Vitchyr H. Pong and Tolly Powell and Alethea Power and Boris Power and Elizabeth Proehl and Raul Puri and Alec Radford and Jack W. Rae and Aditya Ramesh and Cameron Raymond and Francis Real and Kendra Rimbach and Carl Ross and Bob Rotsted and Henri Roussez and Nick Ryder and Mario D. Saltarelli and Ted Sanders and Shibani Santurkar and Girish Sastry and Heather Schmidt and David Schnurr and John Schulman and Daniel Selsam and Kyla Sheppard and Toki Sherbakov and Jessica Shieh and Sarah Shoker and Pranav Shyam and Szymon Sidor and Eric Sigler and Maddie Simens and Jordan Sitkin and Katarina Slama and Ian Sohl and Benjamin Sokolowsky and Yang Song and Natalie M. Staudacher and Felipe Petroski Such and Natalie Summers and Ilya Sutskever and Jie Tang and Nikolas A. Tezak and Madeleine Thompson and Phil Tillet and Amin Tootoonchian and Elizabeth Tseng and Preston Tuggle and Nick Turley and Jerry Tworek and Juan Felipe Cer'on Uribe and Andrea Vallone and Arun Vijayvergiya and Chelsea Voss and Carroll L. Wainwright and Justin Wang and Alvin Wang and Ben Wang and Jonathan Ward and Jason Wei and CJ Weinmann and Akila Welihinda and Peter Welinder and Jiayi Weng and Lilian Weng and Matt Wiethoff and Dave Willner and Clemens Winter and Samuel Wolrich and Hannah Wong and Lauren Workman and Sherwin Wu and Jeff Wu and Michael Wu and Kai Xiao and Tao Xu and Sarah Yoo and Kevin Yu and Qim-ing Yuan and Wojciech Zaremba and Rowan Zellers and Chong Zhang and Marvin Zhang and Shengjia Zhao and Tianhao Zheng and Juntang Zhuang and William Zhuk and Barret Zoph},
	year={2023},
	url={https://api.semanticscholar.org/CorpusID:257532815}
}

@article{yang2025qwen3,
	title={Qwen3 technical report},
	author={Yang, An and Li, Anfeng and Yang, Baosong and Zhang, Beichen and Hui, Binyuan and Zheng, Bo and Yu, Bowen and Gao, Chang and Huang, Chengen and Lv, Chenxu and others},
	journal={arXiv preprint arXiv:2505.09388},
	year={2025}
}

@article{liu2025deepseek,
	title={DeepSeek-V3. 2: Pushing the Frontier of Open Large Language Models},
	author={Liu, Aixin and Mei, Aoxue and Lin, Bangcai and Xue, Bing and Wang, Bingxuan and Xu, Bingzheng and Wu, Bochao and Zhang, Bowei and Lin, Chaofan and Dong, Chen and others},
	journal={arXiv preprint arXiv:2512.02556},
	year={2025}
}

@article{tomilin2025meal,
	title={MEAL: A Benchmark for Continual Multi-Agent Reinforcement Learning},
	author={Tomilin, Tristan and Boogaard, Luka van den and Garcin, Samuel and Grooten, Bram and Fang, Meng and Du, Yali and Pechenizkiy, Mykola},
	journal={arXiv preprint arXiv:2506.14990},
	year={2025}
}

@inproceedings{shridharalfworld,
	title={ALFWorld: Aligning Text and Embodied Environments for Interactive Learning},
	author={Shridhar, Mohit and Yuan, Xingdi and Cote, Marc-Alexandre and Bisk, Yonatan and Trischler, Adam and Hausknecht, Matthew},
	booktitle={International Conference on Learning Representations}
}

@inproceedings{wang2022scienceworld,
	title={ScienceWorld: Is your Agent Smarter than a 5th Grader?},
	author={Wang, Ruoyao and Jansen, Peter and C{\^o}t{\'e}, Marc-Alexandre and Ammanabrolu, Prithviraj},
	booktitle={Proceedings of the 2022 Conference on Empirical Methods in Natural Language Processing},
	pages={11279--11298},
	year={2022}
}

@inproceedings{li2023api,
	title={API-Bank: A Comprehensive Benchmark for Tool-Augmented LLMs},
	author={Li, Minghao and Zhao, Yingxiu and Yu, Bowen and Song, Feifan and Li, Hangyu and Yu, Haiyang and Li, Zhoujun and Huang, Fei and Li, Yongbin},
	booktitle={EMNLP},
	year={2023}
}

@article{yao2024tau,
	title={tau-bench: A Benchmark for Tool-Agent-User Interaction in Real-World Domains},
	author={Yao, Shunyu and Shinn, Noah and Razavi, Pedram and Narasimhan, Karthik},
	journal={arXiv preprint arXiv:2406.12045},
	year={2024}
}

@inproceedings{yang2018hotpotqa,
	title={HotpotQA: A dataset for diverse, explainable multi-hop question answering},
	author={Yang, Zhilin and Qi, Peng and Zhang, Saizheng and Bengio, Yoshua and Cohen, William and Salakhutdinov, Ruslan and Manning, Christopher D},
	booktitle={Proceedings of the 2018 conference on empirical methods in natural language processing},
	pages={2369--2380},
	year={2018}
}

@article{yao2022webshop,
	title={Webshop: Towards scalable real-world web interaction with grounded language agents},
	author={Yao, Shunyu and Chen, Howard and Yang, John and Narasimhan, Karthik},
	journal={Advances in Neural Information Processing Systems},
	volume={35},
	pages={20744--20757},
	year={2022}
}

@article{zhuang2023toolqa,
	title={Toolqa: A dataset for llm question answering with external tools},
	author={Zhuang, Yuchen and Yu, Yue and Wang, Kuan and Sun, Haotian and Zhang, Chao},
	journal={Advances in Neural Information Processing Systems},
	volume={36},
	pages={50117--50143},
	year={2023}
}

@article{deng2023mind2web,
	title={Mind2web: Towards a generalist agent for the web},
	author={Deng, Xiang and Gu, Yu and Zheng, Boyuan and Chen, Shijie and Stevens, Sam and Wang, Boshi and Sun, Huan and Su, Yu},
	journal={Advances in Neural Information Processing Systems},
	volume={36},
	pages={28091--28114},
	year={2023}
}

@inproceedings{liu2018reinforcement,
	title={Reinforcement Learning on Web Interfaces using Workflow-Guided Exploration},
	author={Liu, Evan Zheran and Guu, Kelvin and Pasupat, Panupong and Shi, Tianlin and Liang, Percy},
	booktitle={International Conference on Learning Representations},
	year={2018}
}

@inproceedings{thorne2018fever,
	title={FEVER: a Large-scale Dataset for Fact Extraction and VERification},
	author={Thorne, James and Vlachos, Andreas and Christodoulopoulos, Christos and Mittal, Arpit},
	booktitle={Proceedings of the 2018 Conference of the North American Chapter of the Association for Computational Linguistics: Human Language Technologies, Volume 1 (Long Papers)},
	year={2018},
	organization={Association for Computational Linguistics}
}

@inproceedings{yao2022react,
	title={React: Synergizing reasoning and acting in language models},
	author={Yao, Shunyu and Zhao, Jeffrey and Yu, Dian and Du, Nan and Shafran, Izhak and Narasimhan, Karthik R and Cao, Yuan},
	booktitle={The eleventh international conference on learning representations},
	year={2022}
}

@article{cossu2024continual,
	title={Continual pre-training mitigates forgetting in language and vision},
	author={Cossu, Andrea and Carta, Antonio and Passaro, Lucia and Lomonaco, Vincenzo and Tuytelaars, Tinne and Bacciu, Davide},
	journal={Neural Networks},
	volume={179},
	pages={106492},
	year={2024},
	publisher={Elsevier}
}

@article{gupta2023continual,
	title={Continual pre-training of large language models: How to (re) warm your model?},
	author={Gupta, Kshitij and Th{\'e}rien, Benjamin and Ibrahim, Adam and Richter, Mats L and Anthony, Quentin and Belilovsky, Eugene and Rish, Irina and Lesort, Timoth{\'e}e},
	journal={arXiv preprint arXiv:2308.04014},
	year={2023}
}

@inproceedings{sunlamol,
	title={LAMOL: LAnguage MOdeling for Lifelong Language Learning},
	author={Sun, Fan-Keng and Ho, Cheng-Hao and Lee, Hung-Yi},
	booktitle={International Conference on Learning Representations},
	year={2019}
}

@inproceedings{huang2021continual,
	title={Continual Learning for Text Classification with Information Disentanglement Based Regularization},
	author={Huang, Yufan and Zhang, Yanzhe and Chen, Jiaao and Wang, Xuezhi and Yang, Diyi},
	booktitle={Proceedings of the 2021 Conference of the North American Chapter of the Association for Computational Linguistics: Human Language Technologies},
	pages={2736--2746},
	year={2021}
}

@inproceedings{lin2024mitigating,
	title={Mitigating the alignment tax of rlhf},
	author={Lin, Yong and Lin, Hangyu and Xiong, Wei and Diao, Shizhe and Liu, Jianmeng and Zhang, Jipeng and Pan, Rui and Wang, Haoxiang and Hu, Wenbin and Zhang, Hanning and others},
	booktitle={Proceedings of the 2024 Conference on Empirical Methods in Natural Language Processing},
	pages={580--606},
	year={2024}
}

@article{yu2025memagent,
	title={MemAgent: Reshaping Long-Context LLM with Multi-Conv RL-based Memory Agent},
	author={Yu, Hongli and Chen, Tinghong and Feng, Jiangtao and Chen, Jiangjie and Dai, Weinan and Yu, Qiying and Zhang, Ya-Qin and Ma, Wei-Ying and Liu, Jingjing and Wang, Mingxuan and others},
	journal={arXiv preprint arXiv:2507.02259},
	year={2025}
}

@article{packer2023memgpt,
	title={MemGPT: Towards LLMs as Operating Systems},
	author={Packer, Charles and Wooders, Sarah and Lin, Kevin and Fang, Vivian and Patil, Shishir G and Stoica, Ion and Gonzalez, Joseph E},
	journal={arXiv preprint arXiv:2310.08560},
	year={2023}
}

@article{liu2023think,
	title={Think-in-memory: Recalling and post-thinking enable llms with long-term memory},
	author={Liu, Lei and Yang, Xiaoyan and Shen, Yue and Hu, Binbin and Zhang, Zhiqiang and Gu, Jinjie and Zhang, Guannan},
	journal={arXiv preprint arXiv:2311.08719},
	year={2023}
}

@article{wangvoyager,
	title={Voyager: An Open-Ended Embodied Agent with Large Language Models},
	author={Wang, Guanzhi and Xie, Yuqi and Jiang, Yunfan and Mandlekar, Ajay and Xiao, Chaowei and Zhu, Yuke and Fan, Linxi and Anandkumar, Anima},
	journal={Transactions on Machine Learning Research}
}

@article{chhikara2025mem0,
	title={Mem0: Building production-ready ai agents with scalable long-term memory},
	author={Chhikara, Prateek and Khant, Dev and Aryan, Saket and Singh, Taranjeet and Yadav, Deshraj},
	journal={arXiv preprint arXiv:2504.19413},
	year={2025}
}

@article{rasmussen2025zep,
	title={Zep: a temporal knowledge graph architecture for agent memory},
	author={Rasmussen, Preston and Paliychuk, Pavlo and Beauvais, Travis and Ryan, Jack and Chalef, Daniel},
	journal={arXiv preprint arXiv:2501.13956},
	year={2025}
}

@article{zhang2025survey,
	title={A survey on the memory mechanism of large language model-based agents},
	author={Zhang, Zeyu and Dai, Quanyu and Bo, Xiaohe and Ma, Chen and Li, Rui and Chen, Xu and Zhu, Jieming and Dong, Zhenhua and Wen, Ji-Rong},
	journal={ACM Transactions on Information Systems},
	volume={43},
	number={6},
	pages={1--47},
	year={2025},
	publisher={ACM New York, NY}
}

@article{xi2025rise,
	title={The rise and potential of large language model based agents: A survey},
	author={Xi, Zhiheng and Chen, Wenxiang and Guo, Xin and He, Wei and Ding, Yiwen and Hong, Boyang and Zhang, Ming and Wang, Junzhe and Jin, Senjie and Zhou, Enyu and others},
	journal={Science China Information Sciences},
	volume={68},
	number={2},
	pages={121101},
	year={2025},
	publisher={Springer}
}

@article{jiang2025adaptation,
	title={Adaptation of Agentic AI},
	author={Jiang, Pengcheng and Lin, Jiacheng and Shi, Zhiyi and Wang, Zifeng and He, Luxi and Wu, Yichen and Zhong, Ming and Song, Peiyang and Zhang, Qizheng and Wang, Heng and others},
	journal={arXiv preprint arXiv:2512.16301},
	year={2025}
}

@article{douze2024faiss,
	title={The Faiss library},
	author={Matthijs Douze and Alexandr Guzhva and Chengqi Deng and Jeff Johnson and Gergely Szilvasy and Pierre-Emmanuel Mazaré and Maria Lomeli and Lucas Hosseini and Hervé Jégou},
	year={2024},
	eprint={2401.08281},
	archivePrefix={arXiv},
	primaryClass={cs.LG}
}

@article{Wang2021ASO,
	title={A Survey on Curriculum Learning},
	author={Xin Wang and Yudong Chen and Wenwu Zhu},
	journal={IEEE Transactions on Pattern Analysis and Machine Intelligence},
	year={2021},
	volume={44},
	pages={4555-4576},
	url={https://api.semanticscholar.org/CorpusID:232362223}
}

@inproceedings{zhao-etal-2024-sapt,
	title = "{SAPT}: A Shared Attention Framework for Parameter-Efficient Continual Learning of Large Language Models",
	author = "Zhao, Weixiang  and
	Wang, Shilong  and
	Hu, Yulin  and
	Zhao, Yanyan  and
	Qin, Bing  and
	Zhang, Xuanyu  and
	Yang, Qing  and
	Xu, Dongliang  and
	Che, Wanxiang",
	editor = "Ku, Lun-Wei  and
	Martins, Andre  and
	Srikumar, Vivek",
	booktitle = "Proceedings of the 62nd Annual Meeting of the Association for Computational Linguistics (Volume 1: Long Papers)",
	month = aug,
	year = "2024",
	address = "Bangkok, Thailand",
	publisher = "Association for Computational Linguistics",
	url = "https://aclanthology.org/2024.acl-long.625/",
	doi = "10.18653/v1/2024.acl-long.625",
	pages = "11641--11661"
}

@article{Yang2026AutoSkillEL,
	title={AutoSkill: Experience-Driven Lifelong Learning via Skill Self-Evolution},
	author={Yutao Yang and Junsong Li and Qianjun Pan and Bihao Zhan and Yuxuan Cai and Linge Du and Jie Zhou and Kai Chen and Qin Chen and Xin Li and Bo Zhang and Liang He},
	journal={ArXiv},
	year={2026},
	volume={abs/2603.01145},
	url={https://api.semanticscholar.org/CorpusID:286224498}
}

@book{mardia2000directional,
	title={Directional Statistics},
	author={Mardia, Kanti V and Jupp, Peter E},
	volume={494},
	year={2000},
	publisher={John Wiley \& Sons}
}

@inproceedings{luo-etal-2024-large,
	title = "Large Language Models as Foundations for Next-Gen Dense Retrieval: A Comprehensive Empirical Assessment",
	author = "Luo, Kun  and
	Qin, Minghao  and
	Liu, Zheng  and
	Xiao, Shitao  and
	Zhao, Jun  and
	Liu, Kang",
	editor = "Al-Onaizan, Yaser  and
	Bansal, Mohit  and
	Chen, Yun-Nung",
	booktitle = "Proceedings of the 2024 Conference on Empirical Methods in Natural Language Processing",
	month = nov,
	year = "2024",
	address = "Miami, Florida, USA",
	publisher = "Association for Computational Linguistics",
	url = "https://aclanthology.org/2024.emnlp-main.80/",
	doi = "10.18653/v1/2024.emnlp-main.80",
	pages = "1354--1365",
}

@article{lv2026all,
	title={All-Mem: Agentic Lifelong Memory via Dynamic Topology Evolution},
	author={Lv, Can and Chang, Heng and Tao, Shengyu and Chen, Mingju and Fan, Zhaoxin and Zhang, Ziwei and Guo, Yuchen and Zhou, Shiji},
	journal={arXiv preprint arXiv:2603.19595},
	year={2026}
}

@inproceedings{wang2021minilmv2,
	title={Minilmv2: Multi-head self-attention relation distillation for compressing pretrained transformers},
	author={Wang, Wenhui and Bao, Hangbo and Huang, Shaohan and Dong, Li and Wei, Furu},
	booktitle={Findings of the Association for Computational Linguistics: ACL-IJCNLP 2021},
	pages={2140--2151},
	year={2021}
}

@inproceedings{cheng-etal-2026-mem2evolve,
	title = "{M}em$^2${E}volve: Towards Self-Evolving Agents via Co-Evolutionary Capability Expansion and Experience Distillation",
	author = "Cheng, Zihao  and
	Liu, Zeming  and
	Shan, Yingyu  and
	Wang, Xinyi  and
	Zhu, Xiangrong  and
	Ma, Yunpu  and
	Wang, Hongru  and
	Guo, Yuhang  and
	Lin, Wei  and
	Wang, Yunhong",
	editor = "Liakata, Maria  and
	Moreira, Viviane P.  and
	Zhang, Jiajun  and
	Jurgens, David",
	booktitle = "Proceedings of the 64th Annual Meeting of the {A}ssociation for {C}omputational {L}inguistics (Volume 1: Long Papers)",
	month = jul,
	year = "2026",
	address = "San Diego, California, United States",
	publisher = "Association for Computational Linguistics",
	url = "https://aclanthology.org/2026.acl-long.952/",
	doi = "10.18653/v1/2026.acl-long.952",
	pages = "20784--20831",
	ISBN = "979-8-89176-390-6",
}

@article{shan2026learning,
	title={Learning from Own Solutions: Self-Conditioned Credit Assignment for Reinforcement Learning with Verifiable Rewards},
	author={Shan, Yingyu and Guo, Yuhang and Cheng, Zihao and Liu, Zeming and Zhu, Xiangrong and Wang, Xinyi and Yao, Jiashu and Lin, Wei and Wang, Hongru and Huang, Heyan},
	journal={arXiv preprint arXiv:2606.18810},
	year={2026}
}
		
		\appendix
		
		\section{Details of Datasets and Environments}
		\label{appendix:a}
		\subsection{Embodied Action}
		\textbf{ALFWorld} \cite{shridharalfworld} is an embodied household benchmark built on the ALFRED environment, where agents follow natural language instructions to perform multi-step tasks involving navigation, object manipulation and state changes (e.g., picking up, placing, opening) in simulated indoor scenes. The action space is discrete and text-based, supporting embodied reasoning and planning. We set the maximum episode length to 20 steps, and we report success rate as the evaluation metrics.
		
		\textbf{ScienceWorld} \cite{wang2022scienceworld} is an embodied interactive environment designed to evaluate scientific reasoning and procedural knowledge acquisition through language interaction. The environment simulates virtual laboratory and household settings, where agents follow textual instructions to complete tasks involving multi-step experiments, object manipulation, and causal reasoning (e.g., heating, mixing and measuring). Similar to AlfWorld, its action space is discrete and text-based, consisting of navigation, manipulation, and inquiry actions that allow agents to interact with objects and query the environment. We set its maximum step as 35, and report \textit{Success Rate} as the evaluation metrics. Here we adopt a more stringent evaluation criterion: a task is considered successfully executed only when the environment returns a score of 100. This criterion helps better reflect the correctness of the model’s intermediate steps.
		
		Both two datasets share similar embodied interaction settings with discrete, text-based action spaces, and ScienceWorld covers a wider range of task types about operations in science experiments and provides a more diverse set of actions.
		\subsection{Tool Utilization}
		\textbf{APIBank} \cite{li2023api} is a tool-use benchmark that evaluates an agent’s ability to invoke external APIs to solve user-specified tasks (e.g., retrieving flight information, querying weather, or managing calendar events). Agent first uses 'ToolSearcher' to find suitable tools, call them with appropriate arguments and make next step based on calling results. It focuses on assessing tool utilization, parameter reasoning, and decision making over multi-step API workflows. We select lv1 and lv2 level tasks as our testset, with maximum step as 5.
		
		\textbf{Retail ($\tau$-bench)} \cite{yao2024tau} evaluates agents in real-world vertical domains, including retail and airline services, with a focus on multi-step API tool usage. Agents receive natural language instructions and must execute sequences of tool calls, such as querying inventory, modifying reservations, or calculating prices, across multi-turn interactions with a database. We set its maximum step as 30.
		
		Similar to APIBank, agent in $\tau$-bench acts as assistant, interpreting user instructions and invoking a sequence of APIs to accomplish the requested tasks. This setup evaluates the agent’s ability to plan and execute multi-step operations in a coherent and goal-directed manner.
		
		\subsection{Web Search}
		\textbf{HotpotQA} \cite{yang2018hotpotqa} is a multi-hop question-answering dataset. Following prior approach \cite{zhao2024expel}, agent interacts with Wikipedia API to search for relevant information and generate answers. Its action space includes querying, navigating through pages, and producing answers, enabling evaluation of multi-step reasoning and information retrieval in a structured tool-use context. Following prior approach \cite{zhao2024expel}, \textit{Exact Match} is used to measure the accuracy of the dataset, with maximum step as 7.
		
		\textbf{Webshop} \cite{yao2022webshop} evaluates agents on multi-step online shopping tasks. It contains 1.18 million real-world products collected from Amazon. The environment is implemented using a ChromeDriver-driven browser, where simplified HTML pages serve as observations. Agents interact by entering search queries, clicking buttons, and navigating the page to find and purchase products that satisfy user requests. Once a purchase is finished, the environment provides a reward signaling the degree of match between the acquired product and the user’s specifications. \textit{Task Score} is defined as the mean reward over all tasks, and \textit{Success Rate} represents the fraction of instructions for which $r = 1$, signifying the complete fulfillment of all user criteria. We set its maximum step as 15.
		
		Both two datasets evaluate the ability of agent to search for information on the web according to user requests, select the most relevant results, and make decisions based on the retrieved content.
		
		\subsection{Data Analysis}
		\textbf{LifelongAgentBench} \cite{zheng2025lifelongagentbench}: we select the ``Database'' task from LifelongAgentBench as a component of our training set for data analysis tasks. Within this environment, agents interact with a MySQL database via SQL statements to execute tasks. These tasks encompass 22 SQL-related skills, such as column aliasing, complex filtering with WHERE and HAVING clauses, multi-column grouping, data manipulation (INSERT, UPDATE, DELETE), and nested subqueries.
		
		\textbf{ToolQA} \cite{zhuang2023toolqa} evaluates LLM's ability to use external tools for question-answering. We select \texttt{Tabular Database} task in ToolQA, which requires LLM to perform real-time querying to an external database about coffee prices and compute through a calculator. ToolQA provides two modes of interaction with the database:
		\begin{itemize}
			\item \textit{Custom functions} relies on author-defined functions such as \texttt{LoadDB, GetValue} and \texttt{Filter}, which utilize the pandas library for underlying tabular data queries.
			\item \textit{SQLInterpreter} loads data into SQLite and executes queries by generating SQL statements. 
		\end{itemize}
		Our training set comprises 210 trajectories based on custom functions and 259 based on SQLInterpreter, whereas the test set is entirely SQL-based. Agent is expected to extract common knowledge from the former to achieve superior experience generalization. We set its maximum step as 20.
		
		Both datasets reflect the model's capability to comprehend user requirements and interact with databases through interfaces such as SQL to retrieve relevant data for task completion.
		
		\subsection{Web Browser}
		\textbf{Miniwob++} \cite{liu2018reinforcement} is a web-based interaction benchmark that evaluates agents on a diverse set of tasks simulating basic human–computer interactions, such as clicking buttons, logging in, copying and pasting text, and booking flights. The environment presents agents with simplified web interfaces, where the state is represented by raw HTML code. Agents interact through a discrete action space including clicking elements, moving the mouse, typing text, and selecting options.
		
		\textbf{Mind2web} \cite{deng2023mind2web} is a realistic benchmark for evaluating web agents on open-domain tasks across diverse real-world websites, such as Airbnb and Twitter. It consists of human-annotated demonstrations that require agents to follow high-level user instructions and perform multi-step web navigation. The observation space is the raw HTML of the current webpage, while the action space includes primitive browser interactions such as click, type, and select. Agent performance is evaluated using task-level success rate, measuring whether all steps are correctly executed. We use the cross-task set in Mind2web as our dataset.
		
		Both datasets evaluate an agent’s ability to accomplish diverse tasks through web browsing and interactive page manipulation. Compared to MiniWoB++, Mind2Web features more complex and realistic web environments and more challenging operations, closer to real-world web usage. The clear difference between the two benchmarks provides a natural curriculum, which is beneficial for studying curriculum learning and cross-environment knowledge transfer in web-based agents.
        
		\subsection{Out-of-Distribution Datasets}
		\textbf{FEVER} \cite{thorne2018fever} is a large-scale dataset for fact extraction and verification. As same as HotpotQA, agents in this environment interact with a Wikipedia-based search engine to retrieve evidence for verifying given claims. The task requires the agent to navigate through multiple documents, identify supporting or refuting information, and ultimately classify the claim as \texttt{Supported, Refuted}, or \texttt{NotEnoughInfo}. We evaluate the agent on a subset of 100 test instances, setting the maximum step as 4.
		
		\textbf{Airline ($\tau$-bench)} \cite{yao2024tau} evaluates agents in a complex aviation service context. Crucially, this environment presents a novel domain that was not encountered in our previous training tasks. In this domain, agent interacts with a simulated airline database to manage diverse user requests, such as searching for flights, rebooking itineraries, handling baggage claims, or calculating refund fees. Distinct from the retail task which focuses on inventory and order management, the airline domain introduces more stringent policy constraints and complex business logic, requiring the agent to handle time-sensitive resource allocation and multi-layered compensation rules. We evaluate the agents on a subset of 50 test cases, setting the maximum step as 30 for each interaction.

		\section{Human Annotation in Dataset Construction}
		\label{appendix:b}
		To ensure the quality of the newly generated correct trajectories, we conducted a rigorous human evaluation. We recruited five annotators, all of whom hold at least a Bachelor’s degree in computer science or related fields to ensure their technical proficiency in interpreting agent logs. All participants underwent rigorous training specifically tailored to the diverse task categories and environments involved. They were fully informed of the study's objectives and intended data usage prior to their involvement. All annotators were compensated fairly at a rate of 1.5 times the local minimum wage.
		
		The evaluation assesses the reasonableness of intermediate interaction steps, defined as whether each action represents a logical progression toward the task goal given the current observation. The results confirmed that over 90\% of the intermediate steps in the generated trajectories were reasonable, validating the dataset's suitability for guiding the lifelong learning of agents.
		
		\begin{figure}[t]
			\centering
			\includegraphics[width=\columnwidth]{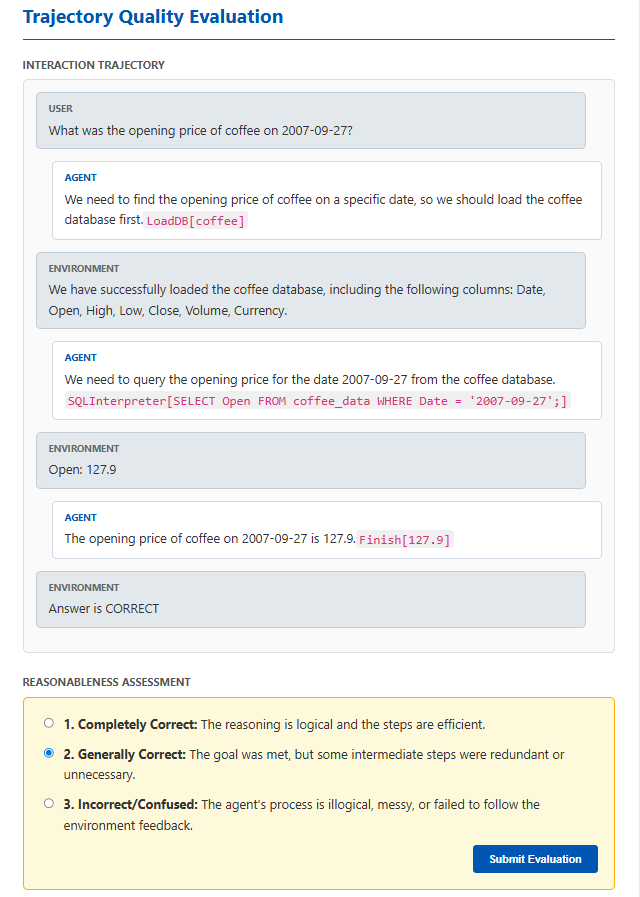}
			\caption{Screen shot of the annotation.}
			\label{annotate}
		\end{figure}

		\begin{table*}[h]
			\resizebox{\textwidth}{!}{%
				\begin{tabular}{cccccccccc}
					\hline
					Train/Eval & AlfWorld & ScienceWorld & APIBank & $\tau$-bench & HotpotQA & Webshop & ToolQA & Miniwob++ & Mind2web \\ \hline
					AlfWorld   & \colorbox{blue!15}{48.51}    &              &          &           &          &         &        &           &          \\
					ScienceWorld   & 50.75    & \colorbox{blue!15}{35.25}        &          &           &          &         &        &           &          \\
					APIbank    &          &              & \colorbox{blue!15}{45.42}    &           &          &         &        &           &          \\
					$\tau$-bench  & 41.79    & 38.13        & 41.06    & \colorbox{blue!15}{26.09}     &          &         &        &           &          \\
					HotpotQA   &          &              &          &           & \colorbox{blue!15}{36.00}       &         &        &           &          \\
					Webshop    & 40.30     & 37.41        & 39.69    & 25.22     & 35.00       & \colorbox{blue!15}{34.00}      &        &           &          \\
					ToolQA     & 45.52    & 34.53        & 35.88    & 23.48     & 31.00       & 32.00      & \colorbox{blue!15}{71.25}  &           &          \\
					Miniwob++  &          &              &          &           &          &         &        & \colorbox{blue!15}{77.08}     &          \\
					Mind2web   & \colorbox{red!20}{44.78}    & \colorbox{red!20}{35.25}        & \colorbox{red!20}{36.64}    & \colorbox{red!20}{25.22}     & \colorbox{red!20}{31.00}       & \colorbox{red!20}{32.00}      & \colorbox{red!20}{68.75}  & \colorbox{red!20}{75.00}        & \colorbox{red!20}{3.97}    \\ \hline
				\end{tabular}%
			}
			\caption{Illustration of the lifelong learning process of GPT-4o-mini using the Synapse. \textsc{Self-learn} denotes the performance of the model when trained and evaluated on a single dataset in isolation.}
			\label{eval}
		\end{table*}
		
		\section{An Intuitive Illustration of Evaluation Metrics}
		\label{appendix:c}
		To facilitate understanding, we present our evaluation metrics: \textit{Overall Performance} and \textit{Backward Transfer}, in an intuitive manner. 
		
		Let $s_{1, \ldots,j}^i$ denote the performance of task $i$ after training on tasks $1, \ldots, j$. As shown in Table~\ref{eval}, the rows below \textsc{AlfWorld} correspond to 9 tasks learned sequentially. After completing training on each task $j$, we evaluate the model on all previously learned tasks $1, \ldots, j$, resulting in a table with a lower-triangular structure.
		For example, after sequentially completed learning on \textsc{AlfWorld} → \textsc{ScienceWorld}, we evaluate the model on both datasets, achieving scores of 50.75 and 35.25, respectively (Row 2).
		
		In addition, the \textsc{Self-learn} row reports the performance $s_i^i$ of the agent when trained and evaluated on a single task in isolation. For example, the second value from the left, 35.97, indicates the model’s performance on \textsc{ScienceWorld} when it is trained exclusively on that task.

		\subsection{Overall Performance}
		
		\textit{Overall Performance} denotes the average performance of the model across all tasks after it has completed learning on the entire task sequence; that is, the mean of all \colorbox{red!20}{values highlighted in red} in the table. This can be formally expressed as:
		\begin{equation}
			\label{eq:example}
			OP = \frac{1}{N}\sum_{i=1}^{N}s_{1, \ldots, N}^i
		\end{equation}
		
		\subsection{Backward Transfer}
		\textit{Backward Transfer} represents the impact of learning subsequent tasks $i, \ldots, N$ to task $i$, which can be formally expressed as:
		\begin{equation}
			\label{bwt}
			BwT = \frac{1}{N-1}\sum_{i=1}^{N-1}\frac{s_{1, \ldots, N}^i-s_{1, \ldots, i}^i}{s_{1, \ldots, i}^i}
		\end{equation}
		This can be intuitively seen by computing the difference between \colorbox{red!20}{values after learning all tasks} and \colorbox{blue!15}{values on corresponding dialogue} in the same column. For example:
		\begin{equation}
			BwT_{miniwob}=(75.00-77.08)/77.08=-2.70\%
		\end{equation}
		indicating that learning \textsc{Mind2web} has a negative effect on \textsc{Miniwob++}.
		
		\section{Investigation on Advanced Retrieval Components}
		\label{retrieval_study}
		\begin{table}[h]
			\centering
			\small
			\resizebox{\columnwidth}{!}{
				\begin{tabular}{lc}
					\toprule
					\textbf{Configuration} & \textbf{Overall Performance} \\ \midrule
					Synapse & 39.18 \\
					Synapse (\textit{+ advanced embedding}) & 40.55 \\
					Synapse (\textit{+ re-ranking}) & 37.29 \\ \midrule
					\textbf{LifeMem} & \textbf{44.13} \\
					\bottomrule
			\end{tabular}}
			\caption{Performance of Synapse equipped with advanced retrieval configuration.}
			\label{tab:retrieval_variants}
		\end{table}
		To address \textbf{RQ4} and verify whether enhanced retrieval capabilities can fundamentally mitigate cross-environment interference, we implement two advanced retrieval variations on top of the Synapse baseline using the \textit{GPT-4o-mini} backbone:
		\begin{itemize}
			\item \textbf{Advanced Embedding:} We substitute the default embedding retriever with a higher-tier model \textit{text-embedding-ada-002}\footnote{https://developers.openai.com/api/docs/models/text-embedding-ada-002}, to evaluate the impact of enhanced semantic matching.
			\item \textbf{Semantic Retrieval + Re-ranking:} To further optimize candidate precision, we utilize \textit{BGE-reranker-v2-m3} \cite{luo-etal-2024-large} to re-rank a broader pool of $5 \times k$ retrieved candidates, subsequently delivering the top-$k$ highest-scoring trajectories to the agent.
		\end{itemize}
		
		As reported in Table~\ref{tab:retrieval_variants}, simply elevating the retrieval components yields marginal or even detrimental effects on agent execution. Upgrading to a stronger embedding model only provides a negligible performance gain, failing to closely approach the superiority of LifeMem. More noticeably, integrating re-ranking mechanism severely degrades the performance. This result may occurs because off-the-shelf, pre-trained re-ranking models are inherently optimized for query-document text relevance in informational retrieval, rather than the intricate behavioral and logic alignment required between sequential agentic trajectories.
		
		These empirical insights firmly demonstrate that simply refining retrieval mechanisms is insufficient to mitigate cross-environment interference, thereby validating the architectural necessity of LifeMem's hierarchical skill abstraction.
		
		\section{Hyperparameter Sensitivity Analysis}
		\label{hyper}
		To evaluate the robustness of LifeMem, we conduct hyper-parameter sensitivity experiments on the AlfWorld using \textit{GPT-4o-mini}. We systematically investigate three key parameters: the retrieval size ($top-k$), the maximum cluster capacity ($L$), and the cluster split threshold coefficient ($\tau_{mrl}$).
		
		\begin{table}[htbp]
			\centering
			\resizebox{\columnwidth}{!}{%
				\begin{tabular}{lcc}
					\toprule
					\textbf{Hyperparameter} & \textbf{Value} & \textbf{Success Rate (\%)} \\
					\midrule
					\multirow{3}{*}{Retrieval Size ($top-k$)} & 2 & 56.72 \\
					& 4 & \textbf{58.21} \\
					& 8 & \textbf{58.21} \\
					\midrule
					\multirow{4}{*}{Cluster Size Limit ($L$)} & 10 & 48.51 \\
					& 30 & 52.99 \\
					& 50 & \textbf{58.21} \\
					& 80 & 51.49 \\
					\midrule
					\multirow{3}{*}{Split Coefficient ($\tau_{mrl}$)} & 0.9 & 54.48 \\
					& 0.95 & \textbf{58.21} \\
					& 1.0 & 57.46 \\
					\bottomrule
				\end{tabular}%
			}
			\caption{Hyperparameter sensitivity analysis.}
			\label{hyperparameters}
		\end{table}
		
		\textbf{Impact of Retrieval Hyperparameter:} $top-k$ controls the number of retrieved trajectories used for in-context learning. We vary $top-k \in \{2, 4, 8\}$ while keeping clustering parameters constant. Results are listed in Table~\ref{hyperparameters}.
		
		The success rate remains stable, demonstrating that the system performance is relatively insensitive to the retrieval volume, indicating that LifeMem relies more on high-quality workflow abstractions than instance quantities.
		
		\textbf{Impact of Workflow Clustering Parameters:} the workflow clustering granularity is governed by the cluster size limit $L$ and the split threshold coefficient $\tau_{mrl}$ (where $\tau_{mrl}=1.0$ means splitting depends solely on $L$). We evaluate $L \in \{10, 30, 50, 80\}$ and $\tau_{mrl} \in \{0.9, 0.95, 1.0\}$, revealing the impact of clustering boundaries:
		\begin{itemize}
			\item \textbf{Under-clustering:} when $L$ is too large or $\tau_{mrl}$ is too low, distinct trajectories merge into single bloated clusters. This under-clustering dilutes the representativeness of the abstracted skills, failing to provide precise guidance and thus degrading performance.
			\item \textbf{Over-clustering:} when $L$ is too small, it may partition data into fragmented sub-clusters too aggressively. This causes LifeMem to overfit to circumstantial noise within specific rollouts rather than forming generalizable workflows, which limits transferability.
		\end{itemize}
		
		\section{Ablation Study}
		\label{ablation}
		\begin{table}[h]
			\resizebox{\columnwidth}{!}{%
				\begin{tabular}{lccc}
					\hline
					& GPT-4o-mini   & Deepseek-v3.2-exp & Qwen3-32b     \\ \hline
					LifeMem                                                           & 44.13         & 48.02             & 46.52         \\
					\begin{tabular}[c]{@{}l@{}}LifeMem\\ \textit{(w/o skills)}\end{tabular}     & 42.84 & 47.09     & 43.88 \\
					\begin{tabular}[c]{@{}l@{}}LifeMem \\ \textit{(w/o examples)}\end{tabular} & 37.61 & 43.46     & 39.65 \\ \hline
				\end{tabular}%
			}
			\caption{Ablation study across three backbone models. \textit{LifeMem (w/o skills)} evaluates performance without the abstracted skills, while \textit{LifeMem (w/o examples)} removes env-aligned trajectories during inference.}
			\label{ablation2}
		\end{table}
		
		\section{Costs in Learning and Inference Time}
		\begin{table}[!htbp]
			\centering
			\small
			\resizebox{\columnwidth}{!}{
				\begin{tabular}{@{\extracolsep{2em}}lcc@{}}
					\toprule
					\textbf{Method} & \textbf{Tokens} & \textbf{Characters} \\
					\midrule
					Synapse   & 1.88k & 11.57k \\
					LifeMem   & 2.18k & 12.72k \\
					ExpeL     & 2.21k & 13.07k \\
					AutoSkill & 4.68k & 28.73k \\
					\bottomrule
				\end{tabular}
			}
			\caption{Comparison of token consumption and length of prompt across methods.}
			\label{cost}
		\end{table}
		In \textbf{memory construction}, LifeMem scales approximately linearly with the number of trajectories (nearly \$1 per 1k trajectories). By utilizing the lightweight model, the overall cost remains moderate and sustainable for long-term lifelong settings.
		
		At \textbf{inference time}, we measured the average context length of the initial agent query across all benchmarks, including the system prompt, few-shot exemplars, abstracted skills, and current task descriptions. As shown in Table~\ref{cost}, token consumption is dominated by system prompts and retrieved trajectories, while LifeMem adds only about 300 tokens over Synapse but yields a 12.63\% relative performance improvement. Under the GPT-4o-mini pricing, this translates to less than \$1 additional inference cost across our whole testset. In contrast, AutoSkill incurs a substantially larger prompt footprint because it lacks workflow-aware organization and therefore injects a bloated global skill profile containing many irrelevant cross-environment behaviors.
		
		\section{Performance in Random Settings}
		\label{random-base}
		\begin{table}[h]
			\centering
			\resizebox{\columnwidth}{!}{
				\begin{tabular}{lcccc}
					\toprule
					& Synapse & ExpeL & AutoSkill & \textbf{LifeMem} \\
					\midrule
					GPT-4o-mini & 39.18 & 39.37 & 32.13 & \textbf{39.63} \\
					Deepseek-v3.2-exp & 41.14 & 40.58 & 41.88 & \textbf{44.34} \\
					\bottomrule
				\end{tabular}
			}
			\caption{Overall performance on random settings.}
			\label{random-setting}
		\end{table}
		
		As shown in Table~\ref{random-setting}, we additionally evaluate LifeMem on the fully random task stream. Although the random task stream is more challenging, as it disrupts opportunities to reuse related experiences, LifeMem consistently remains the best-performing memory approach.
		
		\section{Analysis of Memory Clusters}
		
		\paragraph{Cluster Purity.}
		To examine whether LifeMem organizes trajectories according to consistent execution workflows, we conduct a human annotation study to evaluate cluster purity and analyze the statistical properties of the learned memory clusters. Specifically, three annotators examine the trajectories within each sampled cluster, identify the dominant macro-workflow, and mark trajectories whose execution logic deviates from it. We define Workflow Purity as the proportion of trajectories consistent with the dominant workflow:
		\begin{equation}
		Purity=\frac{\sum_{k=1}^{K} N_{\mathrm{consistent},k}}
{N_{\mathrm{total}}}
		\end{equation}
		
		Across all sampled clusters, LifeMem achieves an average workflow purity of 91\%, providing human evidence that the learned clusters capture consistent execution workflows rather than merely grouping trajectories by embedding similarity.
		
		\begin{figure}[htbp]
			\includegraphics[width=\columnwidth]{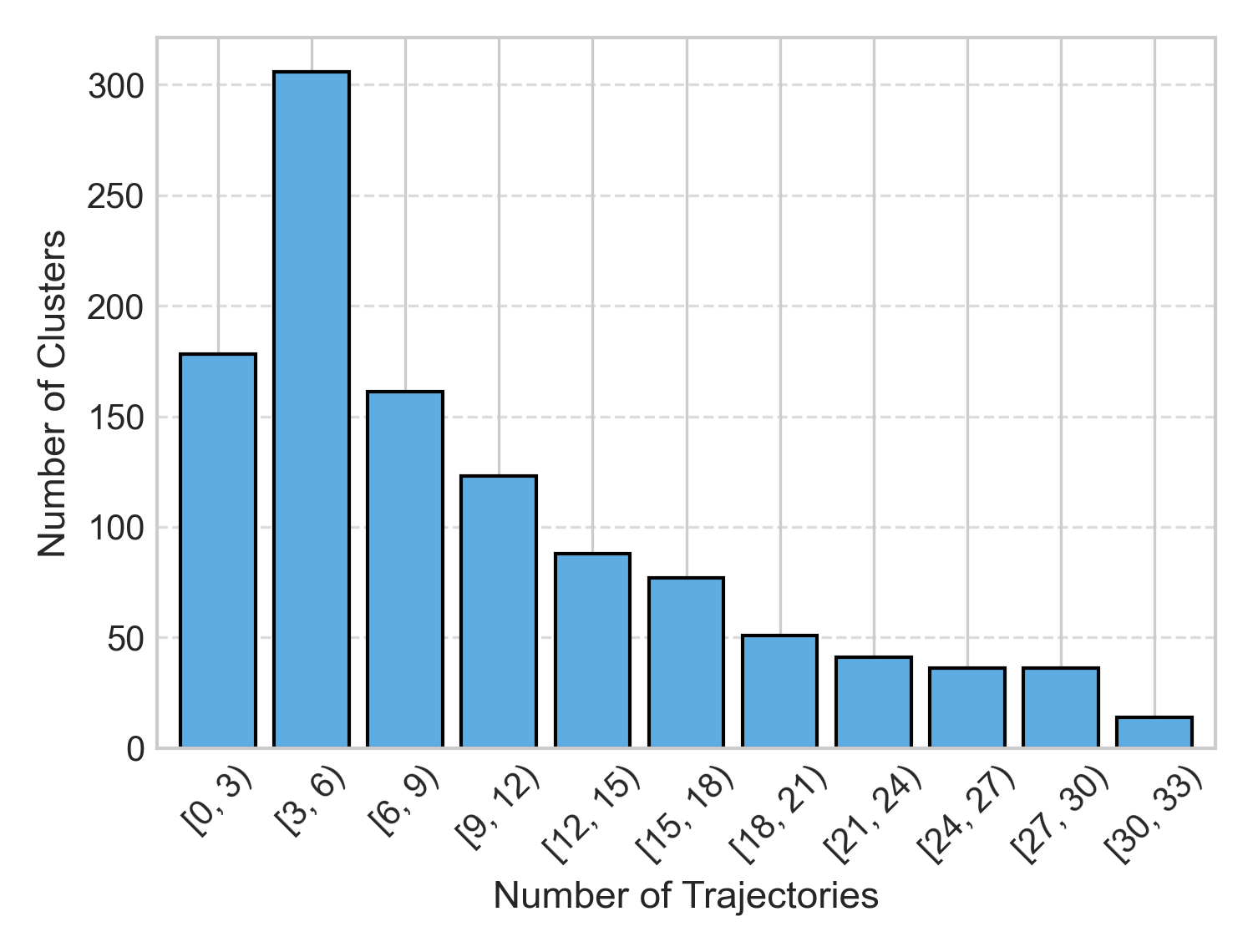}
			\caption{Distribution of the number of trajectories contained in different clusters under the \textbf{similarity} learning order.}
			\label{similarity_cluster}
		\end{figure}
		
		\begin{figure}[htbp]
			\includegraphics[width=\columnwidth]{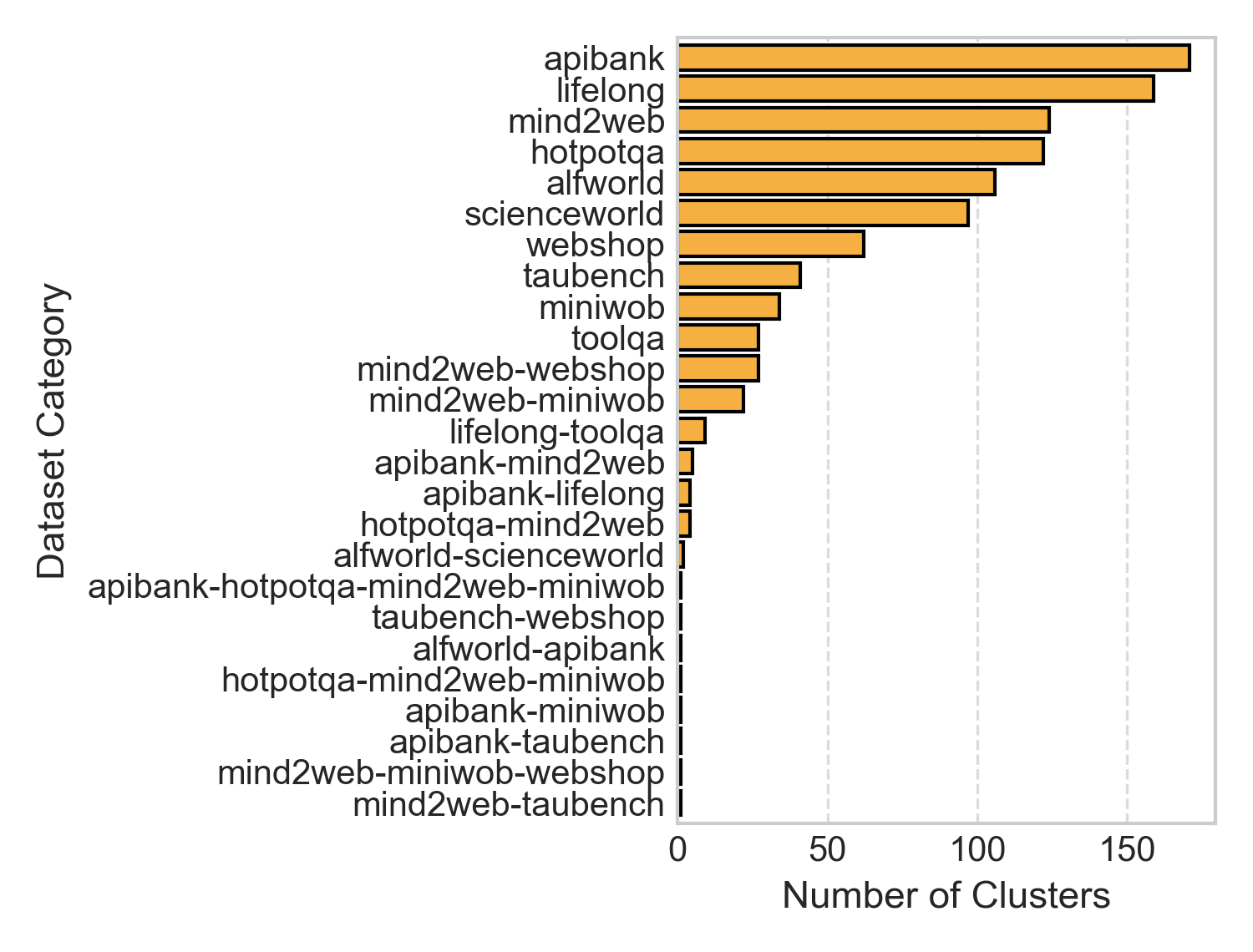}
			\caption{Number of cluster types across datasets, categorized by the datasets to which trajectories within each cluster belong.}
			\label{task_distribution}
		\end{figure}
		
		\paragraph{Adaptive Cluster Granularity.}
		We further analyze the statistical properties of the learned clusters in Figure~\ref{similarity_cluster} and Figure~\ref{task_distribution}. As shown in Figure~\ref{similarity_cluster}, most clusters consolidate multiple trajectories, indicating that LifeMem can abstract recurring execution patterns across tasks instead of storing isolated case-specific experiences.
		
		The cluster statistics also reveal that the learned granularity adapts to the diversity of execution patterns. For environments with relatively repetitive workflows, such as \texttt{WebShop}, clusters contain 29 trajectories on average. In contrast, for more diverse environments such as \texttt{$\tau$-bench}, the average cluster size decreases to 12 trajectories. Consistently, Figure~\ref{task_distribution} shows that datasets with more diverse semantic and action patterns, such as \texttt{APIBank}, are partitioned into a larger number of specialized cluster types, whereas more homogeneous environments, such as \texttt{AlfWorld} and \texttt{ToolQA}, result in fewer and broader clusters. Together, these results suggest that LifeMem can adapt its memory granularity to the structural diversity of trajectories, avoiding both excessive fragmentation and the merging of heterogeneous workflows without manual specification.

		\section{Detailed Numerical Results}
		We present the results of different task training orders on each dataset in Tables~\ref{modelorders}.
		
		\begin{table*}[htbp]
			\resizebox{\textwidth}{!}{%
				\begin{tabular}{lccclccc}
					\hline
					& \multicolumn{3}{c}{\textbf{GPT-4o-mini}} & & \multicolumn{3}{c}{\textbf{Deepseek-v3.2-exp}} \\ \cmidrule(lr){2-4} \cmidrule(lr){6-8}
					\textbf{Dataset} & Similarity & Random & Buffer-based & & Similarity & Random & Buffer-based \\ \hline
					AlfWorld      & 58.21      & 44.03  & 52.24        & & 55.22      & 42.54  & 58.21        \\
					SciWorld      & 33.09      & 23.74  & 23.02        & & 54.68      & 55.40  & 51.80        \\
					APIbank       & 45.42      & 44.27  & 43.89        & & 39.31      & 43.89  & 45.42        \\
					Tau-bench     & 30.43      & 27.83  & 28.70        & & 66.09      & 63.48  & 60.87        \\
					HotpotQA      & 42.00      & 41.00  & 40.00        & & 40.00      & 32.00  & 32.00        \\
					Webshop       & 37.00      & 31.00  & 35.00        & & 40.00      & 39.00  & 40.00        \\
					ToolQA        & 68.75      & 65.00  & 70.00        & & 68.75      & 56.25  & 63.75        \\
					Miniwob++     & 78.72      & 76.60  & 78.72        & & 64.58      & 62.50  & 64.58        \\
					Mind2web      & 3.57       & 3.17   & 3.17         & & 3.57       & 3.97   & 3.97         \\ \hline
					\textbf{Avg.} & \textbf{44.13} & 39.63 & 41.64     & & \textbf{48.02} & 44.34 & 46.73        \\ \hline
				\end{tabular}%
			}
			\caption{Performance of GPT-4o-mini and Deepseek-v3.2-exp with LifeMem under Different Learning Orders.}
			\label{modelorders}
		\end{table*}
		
		We present detailed results of different methods on each dataset in Tables~\ref{all-models-lifelong} of our main results.
		\begin{table*}[htbp]
			\resizebox{\textwidth}{!}{%
				\begin{tabular}{llccccccccc}
					\hline
					& & \multicolumn{5}{c}{\textbf{Overall Performance}}  & \multicolumn{4}{c}{\textbf{Backward Transfer}} \\ \cmidrule(lr){3-7} \cmidrule(lr){8-11}
					\textbf{Model} & \textbf{Dataset} & ReAct & Synapse & ExpeL & AutoSkill & LifeMem        & Synapse   & ExpeL   & AutoSkill & LifeMem         \\ \hline
					
					\multirow{10}{*}{GPT-4o-mini} 
					& AlfWorld                 & 10.45 & 44.78   & 51.49 & 6.72      & 58.21          & -7.69     & -6.75   & -66.65    & +14.70          \\
					& SciWorld                 & 10.79 & 35.25   & 36.69 & 17.99     & 33.09          & 0.00      & +24.37  & +8.70     & +14.98          \\
					& APIbank                  & 44.27 & 36.64   & 41.60 & 46.18     & 45.42          & -19.33    & -6.03   & +1.67     & +6.25           \\
					& Tau-bench                & 26.09 & 25.22   & 32.17 & 22.61     & 30.43          & -3.33     & +2.78   & +4.00     & -5.41           \\
					& HotpotQA                 & 38.00 & 31.00   & 32.00 & 23.00     & 42.00          & -13.89    & +6.67   & -25.81    & +5.00           \\
					& Webshop                  & 27.00 & 32.00   & 35.00 & 36.00     & 37.00          & -5.88     & +2.94   & +12.50    & +8.82           \\
					& ToolQA                   & 73.53 & 68.75   & 65.00 & 45.00     & 68.75          & -3.51     & -1.89   & -2.70     & -1.79           \\
					& Miniwob++                & 27.08 & 75.00   & 79.17 & 97.87     & 78.72          & -2.70     & +2.71   & 0.00      & +2.77           \\
					& Mind2web                 & 2.38  & 3.97    & 3.57  & 3.57      & 3.57           & -         & -       & -         & -               \\ \cline{2-11}
					& \textbf{Avg.}            & 28.84 & 39.18   & 41.85 & 33.22     & \textbf{44.13} & -7.04     & +3.10   & -8.54     & \textbf{+5.66}  \\ \hline
					
					\multirow{10}{*}{Deepseek-v3.2-exp} 
					& AlfWorld                 & 47.01 & 60.45   & 21.64 & 43.28     & 55.22          & -7.95     & +3.54   & -30.13    & -6.34           \\
					& SciWorld                 & 30.94 & 45.32   & 48.92 & 35.25     & 54.68          & +1.61     & +3.03   & -3.92     & +2.70           \\
					& APIbank                  & 43.89 & 43.89   & 44.66 & 43.13     & 39.31          & +2.67     & +1.75   & +2.74     & -7.22           \\
					& Tau-bench                & 54.78 & 63.48   & 63.48 & 62.61     & 66.09          & -3.95     & -3.95   & +7.47     & +4.11           \\
					& HotpotQA                 & 33.00 & 33.00   & 42.00 & 31.00     & 40.00          & 0.00      & 0.00    & -27.91    & +21.21          \\
					& Webshop                  & 27.00 & 39.00   & 37.00 & 23.00     & 40.00          & -2.50     & 0.00    & -25.81    & +5.26           \\
					& ToolQA                   & 37.50 & 63.75   & 71.25 & 61.25     & 68.75          & +2.00     & -3.39   & 0.00      & +1.85           \\
					& Miniwob++                & 93.75 & 64.58   & 62.50 & 62.50     & 64.58          & +3.33     & -3.22   & +4.87     & +3.33           \\
					& Mind2web                 & 2.38  & 3.97    & 3.17  & 3.57      & 3.57           & -         & -       & -         & -               \\ \cline{2-11}
					& \textbf{Avg.}            & 41.14 & 46.38   & 43.85 & 40.62     & \textbf{48.02} & -0.60     & -0.28   & -9.09     & \textbf{+3.11}  \\ \hline
					
					\multirow{10}{*}{Qwen3-32b} 
					& AlfWorld                 & 41.79 & 70.90   & 72.39 & 48.51     & 71.64          & +9.19     & +3.19   & +12.08    & -1.04           \\
					& SciWorld                 & 14.39 & 34.53   & 38.85 & 22.30     & 39.57          & -5.89     & +3.85   & +3.34     & +19.58          \\
					& APIbank                  & 45.04 & 37.02   & 38.93 & 30.54     & 41.22          & -14.92    & -0.97   & -5.86     & +9.08           \\
					& Tau-bench                & 33.04 & 45.22   & 40.87 & 26.09     & 46.96          & +8.34     & -18.96  & -11.77    & 0.00            \\
					& HotpotQA                 & 39.00 & 37.00   & 38.00 & 30.00     & 46.00          & -5.13     & -7.32   & -18.92    & 0.00            \\
					& Webshop                  & 31.00 & 30.00   & 32.00 & 37.00     & 38.00          & 0.00      & +18.52  & +2.78     & 0.00            \\
					& ToolQA                   & 37.50 & 67.50   & 75.00 & 47.50     & 70.00          & -1.82     & 0.00    & -28.36    & +9.80           \\
					& Miniwob++                & 61.70 & 51.06   & 50.00 & 76.60     & 61.70          & -7.70     & -7.70   & -2.69     & +16.00          \\
					& Mind2web                 & 2.38  & 4.37    & 3.57  & 3.57      & 3.57           & -         & -       & -         & -               \\ \cline{2-11}
					& \textbf{Avg.}            & 33.98 & 41.96   & 43.29 & 35.79     & \textbf{46.52} & -2.24     & -1.17   & -6.17     & \textbf{+6.68}  \\ \hline
				\end{tabular}%
			}
		\caption{Detailed results of each dataset on the performance of different models under inter-environment lifelong learning.}
		\label{all-models-lifelong}
		\end{table*}
		
		\section{Examples of the Dataset}
\begin{figure*}[t]
			\centering
			\includegraphics[width=\textwidth]{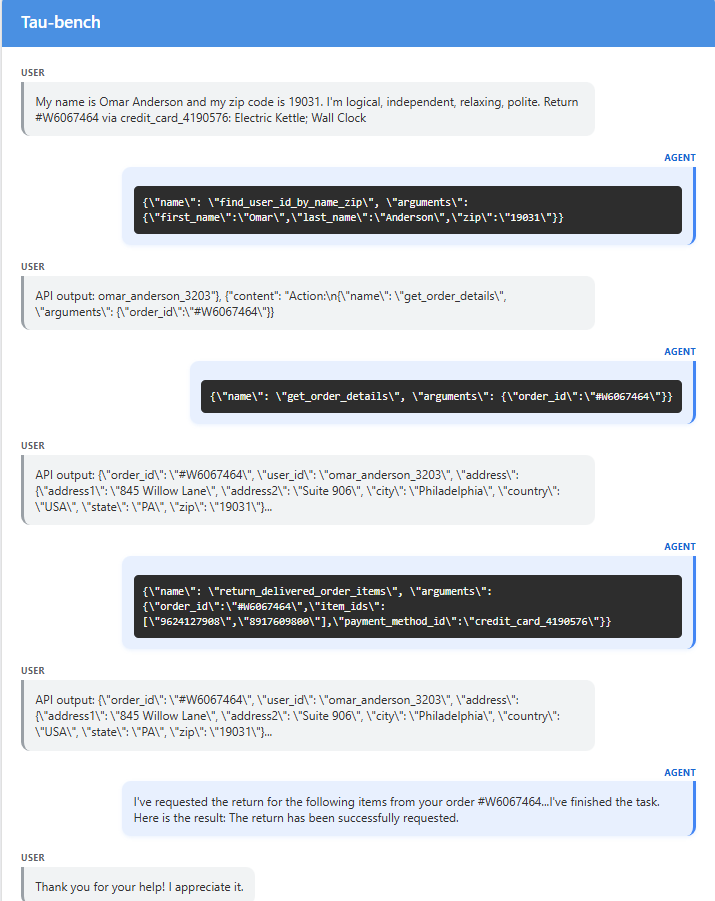}
			\caption{An example of taubench.}
			\label{taubench}
		\end{figure*}

\begin{figure*}[]
			\centering
			\includegraphics[width=\textwidth]{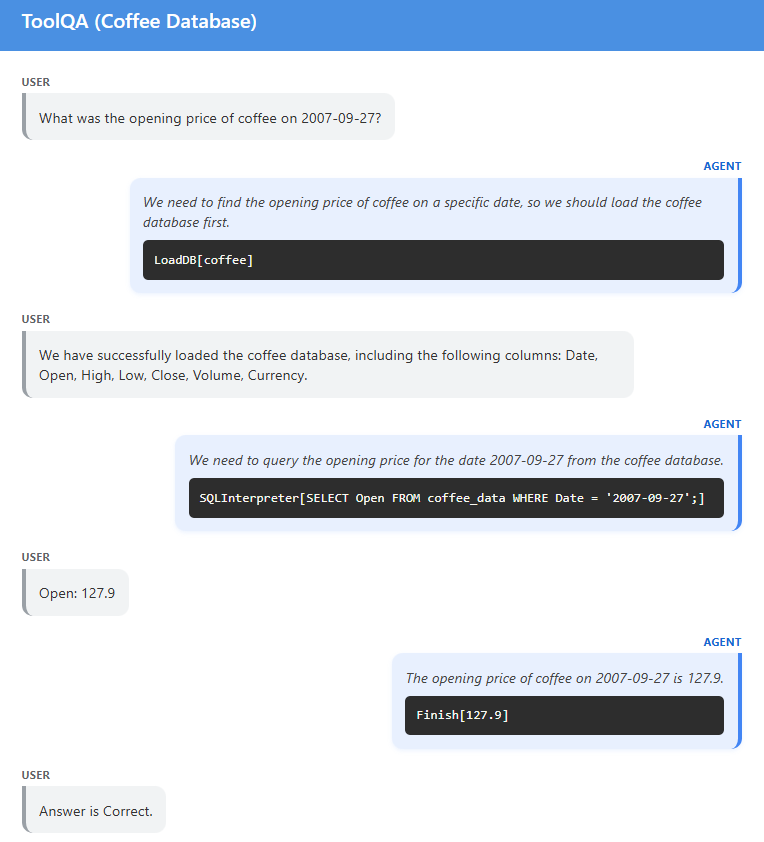}
			\caption{An example of ToolQA.}
			\label{toolqa1}
		\end{figure*}

\begin{figure*}[]
			\centering
			\includegraphics[width=\textwidth]{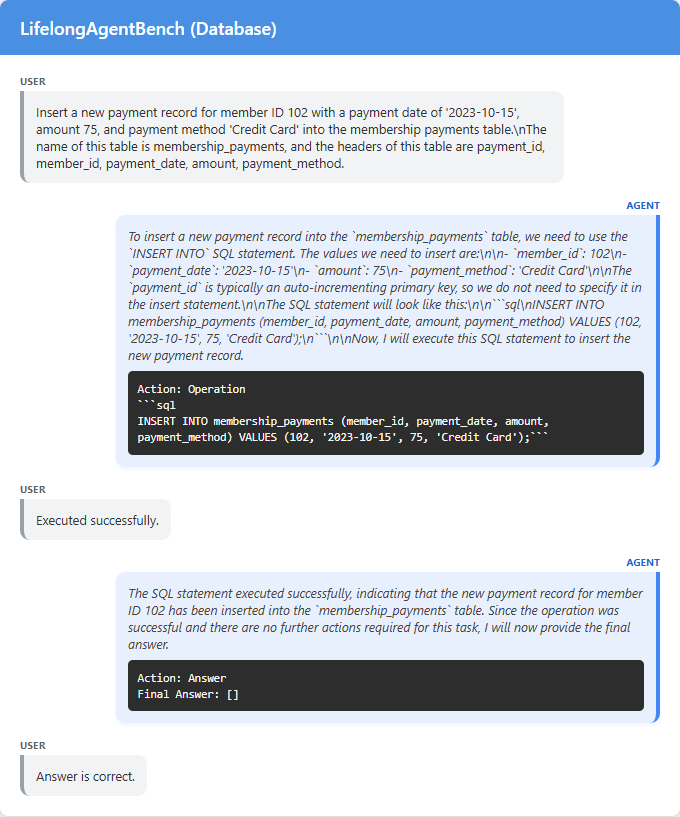}
			\caption{An example of LifelongAgentBench.}
			\label{lifelong}
		\end{figure*}

\begin{figure*}[]
			\centering
			\includegraphics[width=\textwidth]{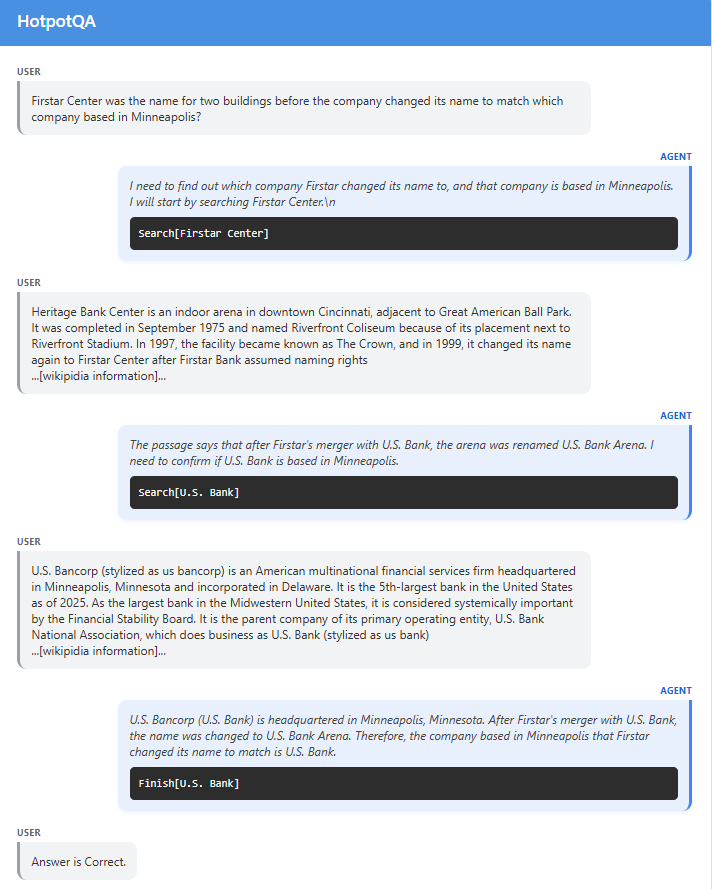}
			\caption{An example of HotpotQA.}
			\label{hotpotqa1}
		\end{figure*}

		We present examples of the four datasets we constructed in Figures~\ref{taubench}, ~\ref{toolqa1}, ~\ref{lifelong}, ~\ref{hotpotqa1}.
		
		\section{Details of Implementation}
		\label{appendix:d}
		We set the temperature of all models to 0.3. For Deepseek-v3.2-exp and Qwen3-32B, we do not enable their reasoning modes.

		\prompttable{Prompt for Categorizing Trajectories into Specific Skill Cluster}{prompts/prompt1.txt}{Prompt for categorizing trajectories into specific skill cluster.}{prompt1}
		\prompttable{Prompt for Cluster Splitting}{prompts/prompt2.txt}{Prompt for cluster splitting.}{prompt2}
		\prompttable{Prompt for Getting Cluster Summary}{prompts/prompt3.txt}{Prompt for getting cluster summary.}{prompt3}
		\prompttable{Prompt for Abstracting skills}{prompts/prompt4.txt}{Prompt for abstracting skills.}{prompt4}
		We present the prompts used in building our agent memory system in Tables~\ref{prompt1}, ~\ref{prompt2}, ~\ref{prompt3}, and~\ref{prompt4}.
		
		\section{Prompts for task execution}
		We present the prompts of each dataset used in inference time of LifeMem in Tables~\ref{alfworld}, ~\ref{scienceworld}, ~\ref{apibank}, ~\ref{retail},  ~\ref{airline},  ~\ref{fever}, ~\ref{hotpotqa}, ~\ref{webshop}, ~\ref{toolqa}, ~\ref{miniwob}, and~\ref{mind2web}.
		
		\prompttable{Prompt for AlfWorld}{prompts/alfworld.txt}{Prompt for AlfWorld dataset.}{alfworld}
		\prompttable{Prompt for ScienceWorld}{prompts/scienceworld.txt}{Prompt for ScienceWorld dataset.}{scienceworld}
		\prompttable{Prompt for APIBank}{prompts/apibank.txt}{Prompt for APIBank dataset.}{apibank}
		\prompttable{Prompt for $\tau$-Bench (Retail)}{prompts/retail.txt}{Prompt for $\tau$-Bench (Retail) dataset. Please refer to \citet{yao2024tau} for the description of the agent policy.}{retail}
		\prompttable{Prompt for $\tau$-Bench (Airline)}{prompts/airline.txt}{Prompt for $\tau$-Bench (Airline) dataset. Please refer to \citet{yao2024tau} for the description of the agent policy.}{airline}
		\prompttable{Prompt for FEVER}{prompts/fever.txt}{Prompt for FEVER dataset.}{fever}
		\prompttable{Prompt for HotpotQA}{prompts/hotpotqa.txt}{Prompt for HotpotQA dataset.}{hotpotqa}
		\prompttable{Prompt for Webshop}{prompts/webshop.txt}{Prompt for Webshop dataset.}{webshop}
		\prompttable{Prompt for ToolQA}{prompts/toolqa.txt}{Prompt for ToolQA dataset.}{toolqa}
		\prompttable{Prompt for Miniwob++}{prompts/miniwob.txt}{Prompt for Miniwob++ dataset.}{miniwob}
		\prompttable{Prompt for Mind2web}{prompts/mind2web.txt}{Prompt for Mind2web dataset.}{mind2web}

	\section{Failure Case Study}
	\label{appendix:h}
	Here, we present a representative failure case study on the \texttt{FEVER} dataset. Since the new task description differs from all previously learned tasks, Synapse may retrieve irrelevant trajectories (e.g., from \texttt{APIBank}). These trajectories exhibit substantial discrepancies in the action space compared to the \texttt{FEVER} environment, thereby misleading the agent’s action generation and ultimately causing task failure. In contrast, our method retrieves environment-aligned trajectories and restricts retrieval to the corresponding \texttt{HotpotQA} repository, effectively avoiding this issue. Moreover, we further provide abstracted insights that offer reusable experience for the model.
	\clearpage

	\begin{table*}[h]
		\centering
		\begin{tcolorbox}[
			width=\textwidth,
			breakable,
			colback=gray!10,
			colframe=gray!75!black,
			title={Failure Case},
			]
			\textbf{\# Task Description}
			\vspace{0.5em}
			
			\textbf{Claim:} The heart beats at a resting rate close to 22 beats per minute.
			
			You are a fact verifier. Solve a question answering task with interleaving \textbf{Thought}, \textbf{Action}, \textbf{Observation} steps. \textbf{Thought} can reason about the current situation, and \textbf{Action} can be three types:
			
			\textbf{Search[entity]}: searches the exact entity on Wikipedia and returns the first paragraph if it exists. If not, it will return some similar entities to search.
			
			\textbf{Lookup[keyword]}: returns the next sentence containing the keyword in the last passage successfully found by \textbf{Search}.
			
			\textbf{Finish[answer]}: returns the answer and finishes the task. The answer should be one of \textbf{REFUTES}, \textbf{SUPPORTS}, or \textbf{NOT ENOUGH INFO}.
			
			You may take a maximum of 4 steps.
			
			\vspace{1em}
			
			\textbf{\# Irrelevant Examples Retrieved from APIBank}
			\vspace{0.5em}
			
			\textbf{User:} I want to add a new exercise to the list for the chest muscle group. The name of the new exercise should be "Chest Fly," and the description should be "Lie down on a bench with your feet flat on the ground and hold a dumbbell in each hand. Extend your arms straight up, then slowly lower them out to the side until your elbows are at shoulder level, and bring them back up." 
			
			\textbf{Assistant:} \textcolor{red}{[ToolSearcher}(keywords='add a new exercise to the list')] 
			
			\textbf{User:} "API: add\_exercise | Description: Add a new exercise to the list of available exercises ..." 
			
			\textbf{Assistant:} \textcolor{red}{[add\_exercise}(name='Chest Fly', description='Lie down on a bench with your feet flat on the ground and hold a dumbbell in each hand. Extend your arms straight up, then slowly lower them out to the side until your elbows are at shoulder level, and bring them back up.', muscle\_group='chest')] 
			
			\textbf{User:} \{"data": \{"status": "success"\}\} 
			
			\textbf{Assistant:} \textcolor{red}{[Finish()]}
			
			\vspace{1em}
			
			\textbf{\# Bad Inference}
			
			\vspace{0.5em}
			
			\textbf{Claim:} The heart beats at a resting rate close to 22 beats per minute. 
			
			\textbf{Action:} \textcolor{red}{[Search}[heart rate]] 
			
			\textbf{Observation:} Invalid Action. Valid Actions are Lookup[$<$topic$>$] Search[$<$topic$>$] and Finish[$<$answer$>$]. 
			
			\textbf{Action:} \textcolor{red}{[Search}[resting heart rate]] 
			
			\textbf{Observation:} Invalid Action. Valid Actions are Lookup[$<$topic$>$] Search[$<$topic$>$] and Finish[$<$answer$>$]. 
		\end{tcolorbox}
		\caption{Failure Case}
		\label{Failure Case}
	\end{table*}

	\begin{table*}[h]
	\centering
	\begin{tcolorbox}[
		width=\textwidth,
		breakable,
		colback=gray!10,
		colframe=gray!75!black,
		title={Success Case},
		]
		\textbf{\# Task Description}
		\vspace{0.5em}
		
		\textbf{Claim:} The heart beats at a resting rate close to 22 beats per minute.
		
		You are a fact verifier. Solve a question answering task with interleaving \textbf{Thought}, \textbf{Action}, \textbf{Observation} steps. \textbf{Thought} can reason about the current situation, and \textbf{Action} can be three types:
		
		\textbf{Search[entity]}: searches the exact entity on Wikipedia and returns the first paragraph if it exists. If not, it will return some similar entities to search.
		
		\textbf{Lookup[keyword]}: returns the next sentence containing the keyword in the last passage successfully found by \textbf{Search}.
		
		\textbf{Finish[answer]}: returns the answer and finishes the task. The answer should be one of \textbf{REFUTES}, \textbf{SUPPORTS}, or \textbf{NOT ENOUGH INFO}.
		
		You may take a maximum of 4 steps.
		
		\vspace{1em}
		
		\textbf{\# Env-align Examples Retrieved from HotpotQA}
		\vspace{0.5em}
		
		\textbf{User:} Swiss music duo Double released their best known single "The Captain of Her Heart" in what year?
		
		\textbf{Assistant:} 
		Search[Double (music duo)]
		
		\textbf{User:} Could not find [Double (music duo)]. Similar: ['Double (band)', ...]
		
		\textbf{Assistant:} 
		Search[Double (band)]
		
		\textbf{User:} Double was a Swiss music duo best known for their hit single "The Captain of Her Heart".
		
		\textbf{Assistant:} 
		Lookup[The Captain of Her Heart]
		
		\textbf{User:} The duo was formed in 1983 in Zürich ...
		
		\textbf{Assistant:} 
		Finish[1986]
		
		\vspace{1em}
		
		\textbf{\# Insights}
		
		\vspace{0.5em}
		
		The following paragraph are insights a teacher agent provided to you. It is MANDATORY for you to follow these insights as CLOSELY as possible as they will help you perform the fact verification tasks efficiently:
		
		In order to successfully complete factual verification tasks, begin by clearly understanding the claim ...
		
		\vspace{1em}
		
		\textbf{\# Success Inference}
		
		\vspace{0.5em}
		
		\textbf{Claim:} The heart beats at a resting rate close to 22 beats per minute.
		
		\textbf{Action:} Search[resting heart rate]
		
		\textbf{Observation:} Heart rate is the frequency of the heartbeat ...

		\textbf{Action:} Search[bradycardia definition]
		
		\textbf{Observation:} Arrhythmias, also known as cardiac arrhythmias, are irregularities ...
		
		\textbf{Action:} Finish[REFUTES]
		
		\textbf{Observation:} Answer is CORRECT
		
	\end{tcolorbox}
	\caption{Success Case}
	\label{Success Case}
	\end{table*}

    \clearpage

	\end{document}